\documentclass{article}
\makeatletter\def\input@path{{template/}}\makeatother
\usepackage[T1]{fontenc}

\usepackage{iclr2027_conference,times}
\iclrfinalcopy
\renewcommand{\headrulewidth}{0pt}

\usepackage{amsmath,amsfonts,bm}

\def\eqref#1{equation~\ref{#1}}

\def\1{\bm{1}}

\DeclareMathAlphabet{\mathsfit}{\encodingdefault}{\sfdefault}{m}{sl}
\SetMathAlphabet{\mathsfit}{bold}{\encodingdefault}{\sfdefault}{bx}{n}

\usepackage{booktabs}
\usepackage{graphicx}
\usepackage{microtype}
\usepackage{tabularx}
\usepackage{ragged2e}
\newcolumntype{L}[1]{>{\RaggedRight\arraybackslash\hyphenpenalty=10000\exhyphenpenalty=10000}p{#1}}
\usepackage{longtable}
\usepackage{float}
\usepackage{placeins}
\usepackage{etoolbox}
\AtBeginEnvironment{table}{\setlength{\belowcaptionskip}{10pt}}
\usepackage{xcolor}
\usepackage{colortbl}
\definecolor{rankink}{HTML}{000000}
\definecolor{rankwash}{HTML}{F0F4F8}
\definecolor{activationink}{HTML}{000000}
\definecolor{activationwash}{HTML}{F0F5F2}
\definecolor{steeringink}{HTML}{000000}
\definecolor{steeringwash}{HTML}{F8F3ED}
\definecolor{overallink}{HTML}{000000}
\definecolor{overallwash}{HTML}{F3F3F3}
\definecolor{tablewash}{HTML}{F5F5F5}
\definecolor{tablerule}{HTML}{000000}
\newcommand{\agentlogo}[1]{\raisebox{-.12em}{\includegraphics[width=.9em,height=.9em,keepaspectratio]{figures/logos/#1.pdf}}\hspace{.35em}}

\definecolor{takeawaybg}{HTML}{F5F6F1}
\definecolor{takeawayborder}{HTML}{A4AB97}
\definecolor{takeawayink}{HTML}{596448}
\newcounter{takeaway}
\newcommand{\takeaway}[2]{%
  \par\addvspace{5pt}%
  \refstepcounter{takeaway}%
  \noindent\begingroup
  \setlength{\fboxsep}{6pt}\setlength{\fboxrule}{0.45pt}%
  \fcolorbox{takeawayborder}{takeawaybg}{%
    \begin{minipage}{\dimexpr\linewidth-2\fboxsep-2\fboxrule\relax}
    \normalsize\RaggedRight
    \textbf{\textcolor{takeawayink}{Takeaway \thetakeaway.}\ #1}\par
    \setlength{\parskip}{2pt}#2
    \end{minipage}}%
  \endgroup\par\addvspace{5pt}%
}

\usepackage{tcolorbox}
\definecolor{promptblue}{HTML}{385B79}
\definecolor{promptbluewash}{HTML}{F3F7FB}
\definecolor{promptteal}{HTML}{366C64}
\definecolor{prompttealwash}{HTML}{F3F8F6}
\newtcolorbox{judgeprompt}[3]{%
  colback=#2,colframe=#1!55!white,colbacktitle=#1!13!white,
  coltitle=#1,fonttitle=\bfseries\normalsize,title=#3,
  fontupper=\small,boxrule=.45pt,titlerule=.35pt,arc=1.2pt,
  left=9pt,right=9pt,top=7pt,bottom=7pt,
  toptitle=4pt,bottomtitle=4pt,boxsep=0pt,
  before skip=8pt,after skip=7pt,
  before upper={\raggedright\setlength{\parskip}{4pt}}}
\newcommand{\promptfield}[1]{\textcolor{promptblue}{\texttt{<#1>}}}

\usepackage{hyperref}
\hypersetup{hidelinks}
\usepackage{url}

\newcommand{\bench}{\textsc{SAEScientist-Bench}}

\definecolor{tableheader}{HTML}{F3EFE8}
\definecolor{casegray}{HTML}{F1F3F5}
\definecolor{caseteal}{HTML}{267D83}
\definecolor{casebrown}{HTML}{BB7449}
\definecolor{casegrayink}{HTML}{526272}
\definecolor{caseblue}{HTML}{426A98}

\newcommand{\caseshift}[1]{\textcolor{casebrown}{\textbf{#1}}}
\newcommand{\casekeep}[1]{\textcolor{caseblue}{\textbf{#1}}}

\graphicspath{{figures/}}

\begin{document}
\thispagestyle{plain}
\raggedbottom
\brokenpenalty=10000
\title{\textsc{SAEScientist-Bench}: Can AI Agents Conduct\\Autonomous SAE Interpretability Research?}
\author{Yuqiao Tan\textsuperscript{1,2}, Shizhu He\textsuperscript{1,2}\thanks{\enskip Corresponding author.}, Jun Zhao\textsuperscript{1,2}, Kang Liu\textsuperscript{1,2} \\
\textsuperscript{1}The Key Laboratory of Cognitive Intelligence, Institute of Automation, CAS \\
\textsuperscript{2}School of Artificial Intelligence, University of Chinese Academy of Sciences \\
\texttt{tanyuqiao2025@ia.ac.cn}, \texttt{\{shizhu.he,jzhao,kliu\}@nlpr.ia.ac.cn}
}

\maketitle

\begin{abstract}
While research on recursive self-improvement (RSI) has predominantly automated model training pipelines, reliable autonomous development demands a missing pillar: post-hoc monitoring and auditing to understand what models learn and ensure safe alignment. Mechanistic interpretability tools are essential to bridge this gap, among which Sparse Autoencoders (SAEs) serve as a cornerstone by isolating interpretable features for model inspection and steering. In this paper, we introduce \bench{} to evaluate whether AI agents can act as scientists utilizing SAE tools for autonomous mechanistic discovery. Given a target concept, an agent designs contrastive probes and navigates a Gemma Scope dictionary of 131K+ features in Gemma-2-9B-IT to discover the optimal feature, evaluated against curated expert reference features anchored on Neuronpedia across activation rank, concept selectivity on contrastive texts, and causal steering.
Across 10 agent configurations and 20 tasks, frontier agents demonstrate genuine discovery capabilities and lead different evaluation dimensions, but remain well behind the expert baseline, approaching expert levels on separating target concepts from contrastive controls while lagging substantially in causal generation steering. Further analysis reveals that although agents can design contrasts to rule out spurious candidates, they frequently misinterpret experimental measurements. These results establish experimental model understanding as a measurable capability for closed-loop autonomous AI R\&D. Our code is available at \url{https://github.com/Trae1ounG/SAEScientist}.
\end{abstract}

\begin{figure}[h]
    \centering
    \includegraphics[width=\linewidth]{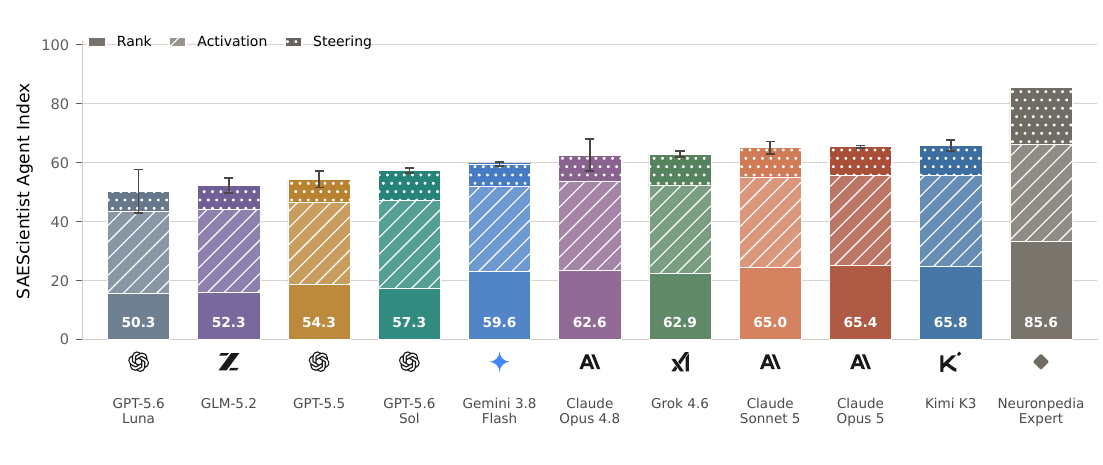}
    \caption{\textbf{SAEScientist Agent Index.} Agents are ordered by their composite Overall score, with each bar decomposed into equally weighted \textit{Rank, Activation, and Steering} components. Error bars denote standard deviations across three independent runs.
    }
    \label{fig:agent-index}
\end{figure}

\section{Introduction}

Recursive self-improvement (RSI) motivates the development of agents that can contribute to successive stages of AI research and development \citep{mahmoud2026rsibench,meng2026rsibenchdata}. Current systems and benchmarks study automated experimentation, agent modification, machine learning engineering, paper reproduction, and post-training \citep{lu2024aiscientist,zhang2025darwin,chan2024mlebench,starace2025paperbench,rank2026posttrainbench,tan2026posttrainbench0}. Understanding the resulting models is another part of this research process. As agents take on more development work, we also need to measure their ability to investigate what trained models represent and how interventions affect their behavior.

Model evaluation and alignment research motivate this need. Reward hacking formalizes how optimizing a proxy objective can conflict with the intended objective \citep{skalse2022defining}. Experiments with deliberately trained deceptive behaviors further show that such behaviors can persist through subsequent safety training \citep{hubinger2024sleeper}. Interpretability and auditing agents offer tools for investigating model representations and hidden behaviors \citep{shaham2024maia,bricken2025auditing}. Evaluating their experimental skills can help establish how reliably agents perform this part of model analysis.

Sparse autoencoders (SAEs) provide a practical tool for these investigations. They decompose model activations into sparse combinations of learned feature directions, many of which correspond to recognizable concepts \citep{cunningham2023sparse,bricken2023monosemanticity,gao2024scaling,templeton2024monosemanticity}. Researchers can examine the texts that activate a feature and intervene along its direction to test effects on model outputs \citep{marks2024circuits,arad-etal-2025-saes}. Steering performance depends on feature selection, and interpretability alone is an insufficient proxy for steering utility \citep{wu2025axbench,wang2026utility}. Pretrained dictionaries such as Gemma Scope make these experiments accessible without training a new SAE \citep{lieberum2024gemmascope}.

Recent agents explain SAE features, discover candidate features, and analyze circuits through interaction with interpretability tools \citep{han2026sage,marinllobet2026automated,khan2026circuitexplainers}. These advances motivate a common evaluation of how well different agents carry out feature discovery. An agent must design informative probes, compare candidates, and select a feature whose behavior holds up under evaluation. Measuring both activation behavior and steering outcomes makes these abilities observable and captures distinctions that either aspect alone can miss \citep{wu2025axbench,wang2026utility}.

To address this challenge, we introduce \bench{}, a benchmark designed to evaluate AI agents as scientists utilizing SAE tools for mechanistic discovery. Covering 20 tasks across diverse concept domains (e.g., multilingual understanding, specialized formats, and safety-critical topics), an agent designs contrastive probes to navigate pretrained SAE dictionaries in Gemma-2-9B-IT \citep{gemma2team} and selects optimal features. Discovered features are systematically evaluated against curated expert reference baselines anchored on Neuronpedia features \citep{lin2023neuronpedia} under a unified scoring framework comprising \textit{\textbf{activation rank}}, \textit{\textbf{activation selectivity}}, and \textit{\textbf{causal steering}}. 

Across 10 frontier agents and these tasks, empirical evaluations demonstrate that agents exhibit genuine discovery capabilities, with Kimi K3 \citep{kimi2026k3} achieving the highest composite overall score, followed closely by Claude Opus 5 and Sonnet 5, while GPT-5.6 Sol and Grok 4.6 demonstrate strong reasoning and steering capabilities. Different frontier models excel in distinct dimensions: Opus leads in activation rank, Kimi leads in activation selectivity, and Grok 4.6 leads in steering efficacy. However, a substantial gap remains compared to the expert reference baseline. While top agents approach expert levels on activation selectivity, reaching 92.91 against the Expert baseline of 98.92, their ability to causally steer model generation falls markedly short, scoring only 31.47 against 57.75. Further qualitative and behavioral analysis indicates that agents often misread experimental measurements, struggle to separate format from concept, and produce features that degrade downstream generation. Our contributions are as follows:
\begin{itemize}
\setlength{\itemsep}{3pt}
\setlength{\parskip}{0pt}
\item We introduce \bench{}, comprising 20 discovery tasks across diverse concept domains, where agents design contrastive probes and navigate a 131K+ feature dictionary in Gemma-2-9B-IT to discover SAE features.
\item We establish a standardized evaluation framework measuring activation rank, activation selectivity, and causal steering against Neuronpedia expert reference baselines, evaluating 10 frontier agents across multiple independent runs.
\item We provide systematic behavioral analyses of authored probes, candidate comparisons, and steering outputs, revealing how agents interpret experimental evidence and where autonomous scientific discovery currently succeeds and fails.
\end{itemize}

\begin{figure}[t]
    \centering
    \includegraphics[width=\linewidth]{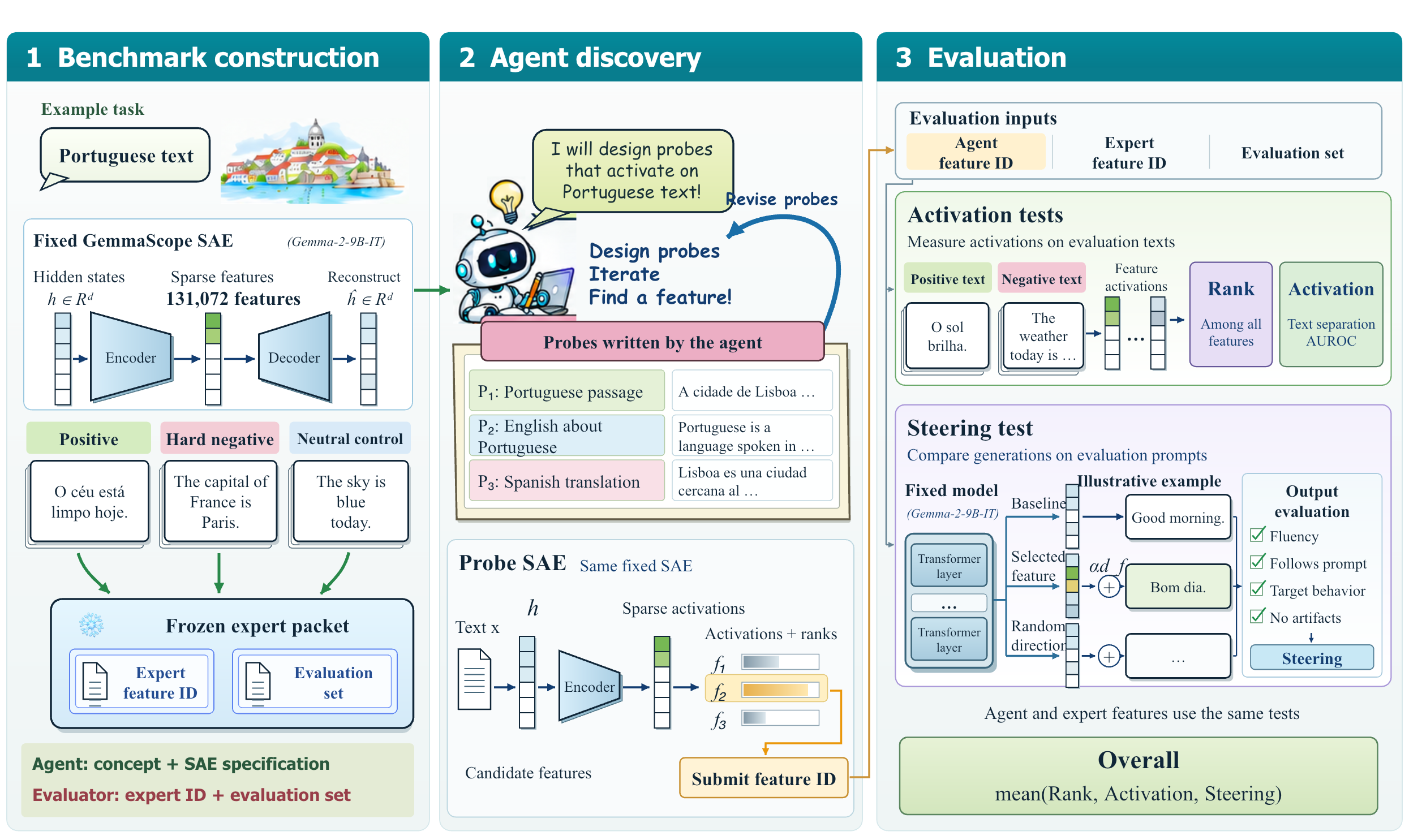}
    \caption{\textbf{SAEScientist-Bench workflow.} For a concept and fixed SAE, the agent writes probes $P_i$, compares candidates $f_j$, and submits a feature. The evaluator measures Rank, Activation, and Steering using Expert and the evaluation set. Portuguese texts and replies illustrate the process.}
    \label{fig:pipeline}
\end{figure}

\section{Preliminaries: SAE Features and Steering}
\label{sec:sae-preliminaries}

\paragraph{SAE formulation and feature activations.} A language model maps input text into token-level hidden states. At a chosen layer, a Sparse Autoencoder (SAE) encodes a hidden state $h \in \mathbb{R}^d$ into a sparse vector of nonnegative feature activations $z \in \mathbb{R}^m$, reconstructing the state via a linear decoder:
\begin{equation}
z = \operatorname{Encoder}(h), \qquad \hat{h} = b_{\rm dec} + \sum_{f=0}^{m-1} z_f d_f,
\label{eq:sae-reconstruction}
\end{equation}
where $m$ is the dictionary size, $b_{\rm dec}$ is a reconstruction bias, and $d_f = W_{\rm dec}[f,:]$ denotes feature $f$'s decoder direction. SAE training balances reconstruction fidelity with sparsity \citep{cunningham2023sparse}. In Gemma Scope \citep{lieberum2024gemmascope}, JumpReLU applies learned activation thresholds to ensure that only a small subset of features activate per token \citep{rajamanoharan2024jumping}. Each coordinate $f$ corresponds to an isolated latent feature, and its scalar activation $z_f$ quantifies the expression of that feature on the input text.

\paragraph{Feature steering via activation addition.} Beyond measuring activations, a feature can be used to causally intervene on model generation. Specifically, activation steering adds the feature's decoder direction $d_f$ directly to the model's hidden state during inference:
\begin{equation}
h \leftarrow h + \alpha d_f,
\label{eq:sae-steering}
\end{equation}
where $\alpha$ controls the intervention strength. The language model then continues generation from this modified state. By comparing model outputs before and after intervention, one can directly evaluate the causal effect of feature $f$ on downstream behavior \citep{arad-etal-2025-saes}.

\section{SAEScientist-Bench}
\label{sec:benchmark}

\begin{figure}[!t]
    \centering
    \includegraphics[width=\linewidth]{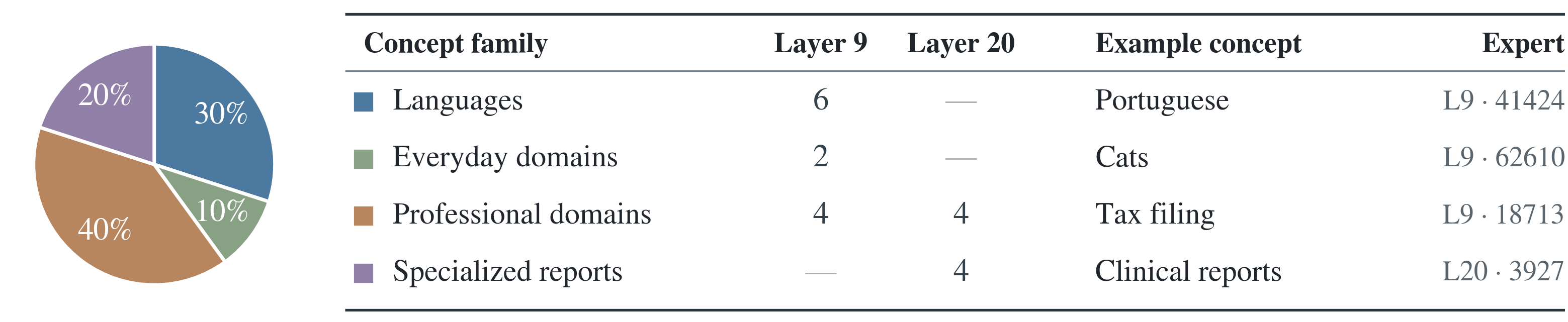}
    \caption{\textbf{Benchmark coverage.} Category shares and task counts by SAE layer, with an example concept and Expert feature for each category. All tasks have equal weight. Appendix~\ref{app:task-inventory} lists the full set.}
    \label{tab:task-design}
    \end{figure}

\subsection{Task Construction}

A task pairs a target concept with a specific layer of the base language model, Gemma-2-9B-IT \citep{gemma2team}, equipped with pretrained Gemma Scope residual-stream SAEs \citep{lieberum2024gemmascope}. The agent is tasked with discovering a single feature within that layer's SAE dictionary (131,072 features) that best represents the concept. As illustrated in Figure~\ref{tab:task-design}, \bench{} comprises 20 discovery tasks across layers 9 and 20, covering diverse concept domains including multilingual understanding (e.g., Portuguese, Spanish, Latin, Turkish), specialized document formats (e.g., earnings reports, tax filing, job postings), and domain-specific knowledge (e.g., clinical symptom reports, pharmaceutical dosing).

For each task, the benchmark provides an evaluation suite consisting of three types of text: positive texts expressing the intended concept, hard-negative texts presenting confusable alternatives (e.g., discussing Portuguese in English), and neutral texts providing unrelated controls. In addition, each task includes evaluation prompts to test causal steering effects on downstream generation. To establish a rigorous reference standard, each task is paired with an Expert reference feature anchored in Neuronpedia's feature repository \citep{lin2023neuronpedia}, established either directly from public steering presets (e.g., Cat) or through standard expert curation workflows across positive and contrastive texts. These task descriptions, Expert features, and evaluation suites remain strictly frozen during agent discovery and are reserved solely for post-submission benchmarking.

\subsection{Interaction Protocol}

The agent receives the concept description, base-model and SAE identifiers, hook point, and dictionary width. Expert and the evaluation set remain reserved for evaluation. The \texttt{probe\_sae} interface accepts up to 64 agent-written texts per request and either retrieves the top-$k$ activating features or measures a supplied set of candidates. Responses include activations and ranks in the full dictionary. The model and its execution harness jointly carry out this investigation. The agent revises its texts and candidates, then submits one feature ID for the evaluator to test through activation and steering.

In Figure~\ref{fig:pipeline}, $P_i$ denotes an agent-written probe and $f_j$ a candidate feature. The subscripts enumerate the illustrated probes and candidates. We use $f$ for the submitted feature and $f_{\rm exp}$ for Expert. Agent-written probes guide discovery, while the separate evaluation set measures the final submission. During discovery, access is restricted to the probe interface and the provided workspace. The same fixed base model and SAE support both discovery measurements and subsequent evaluation.

\subsection{Measurements and Scores}
\label{sec:score-scale}

We evaluate each submitted feature across three complementary dimensions, which are subsequently averaged across tasks: \textbf{Activation Rank} evaluates how prominently a feature activates relative to the dictionary, \textbf{Activation Selectivity} evaluates concept separation between positive texts and contrastive (negative and neutral) controls, and \textbf{Causal Steering} measures downstream generation change under feature intervention.

\paragraph{Activation Rank ($\mathrm{Rank}$).} An effective SAE feature should be prominently activated when the model processes its target concept, rather than being overshadowed by irrelevant dictionary directions. To assess whether an agent identifies a sufficiently prominent feature for the concept, we measure its activation rank relative to the expert reference baseline across positive evaluation texts. For each positive text, a feature's text-level activation is computed as the mean of its three largest non-special-token activations across the sequence, providing a more stable estimate than single-token maximums (Section~\ref{sec:behavior}). We then determine the feature's rank against the full dictionary, where higher activation corresponds to a lower numerical rank and inactive features receive the worst possible rank equal to the dictionary size. Let $r_f$ and $r_{\rm exp}$ denote the average dictionary ranks of the submitted feature and the Expert baseline over positive texts. We compute the relative rank score scaled to a 100-point reference:
\begin{equation}
\mathrm{Rank}=100 \times \frac{2\,r_{\rm exp}}{r_f+r_{\rm exp}}.
\label{eq:rank-current}
\end{equation}
This score is defined in $[0, 200]$, where a score of 100.0 indicates parity with the expert baseline, values above 100.0 indicate a feature ranking ahead of expert, and values below 100.0 indicate lower prominence.

\paragraph{Activation Selectivity ($\mathrm{Activation}$).} Beyond raw activation strength on target texts, a monosemantic feature should respond selectively to the target concept itself rather than to confusable or spurious patterns. We assess this on the evaluation suite by computing the AUROC separating positive texts from contrastive controls (pooling hard-negative and neutral texts), crediting half a point for ties, and scaling the result to $[0, 100]$:
\begin{equation}
\mathrm{Activation}=100 \times \max(0,2\,\mathrm{AUROC}-1).
\label{eq:activation-current}
\end{equation}
The score lies in $[0, 100]$, where 100.0 denotes complete separation of positive texts from contrastive controls, and 0 indicates chance-level or reversed discrimination.

\paragraph{Causal Steering ($\mathrm{Steering}$).} Feature discovery is ultimately validated by its ability to causally steer model generation toward the target concept. On each evaluation prompt, the base model generates completions under three conditions: unmodified baseline inference, feature steering ($h \leftarrow h + \alpha d_f$), and a norm-matched random-direction control. An automated judge rates target relevance and instruction preservation on a 0--4 scale, while separately flagging degeneration. Let $T_f$, $T_{\rm base}$, and $T_{\rm random}$ denote the average target-relevance ratings across prompts and judge passes. The steering score measures the net increase in target expression beyond the stronger control condition, scaled to $[0, 100]$:
\begin{equation}
\mathrm{Steering}=100 \times \max\left(0,\frac{T_f-\max(T_{\rm base},T_{\rm random})}{4}\right).
\label{eq:steering-current}
\end{equation}
The score lies in $[0, 100]$, with higher values indicating stronger causal induction of target expressions. We also denote this net gain as Target Effect ($\Delta_f=\mathrm{Steering}/100$). Crucially, $\mathrm{Steering}$ isolates the magnitude of induced target expression; instruction preservation and output degeneration capture complementary dimensions of generation quality and are evaluated alongside the primary score. For alternative features, the intervention strength $\alpha$ is calibrated on five held-out prompts prior to final evaluation to satisfy a minimum non-degeneration threshold, while Expert retains its frozen reference scale (Appendix~\ref{app:steering-protocol}).

\paragraph{Overall Score.} As a default summary index to present overall discovery performance, each task's overall score is computed as the unweighted arithmetic mean across the three dimension scores:
\begin{equation}
\mathrm{Overall}=\frac{\mathrm{Rank}+\mathrm{Activation}+\mathrm{Steering}}{3}.
\label{eq:overall-score}
\end{equation}
The Expert baseline achieves an overall score of 85.56 under this default setting. While this composite provides a unified view for leaderboard presentation, individual metrics reflect complementary dimensions of feature quality and are evaluated independently (Table~\ref{tab:agent-scores-ranked} and Appendix~\ref{app:metric-dictionary}). All reported benchmark scores represent this unweighted mean across all 20 tasks.

\section{Experimental Setup}

\subsection{Evaluated Models and Harnesses}
We evaluate 10 representative frontier agent configurations across our 20 discovery tasks. The evaluated models span major model families: Kimi K3 \citep{kimi2026k3}, Claude Opus 5, Claude Sonnet 5, Claude Opus 4.8, Grok 4.6, Gemini 3.8 Flash, GLM-5.2, and the OpenAI series (GPT-5.6 Sol, GPT-5.5, GPT-5.6 Luna). Sol and Luna operate within the Codex harness, while all other models are deployed in Cursor. To ensure parity and prevent data contamination, all agents receive identical task instructions and interaction interfaces, with external network access disabled. Models operate under unconstrained interactive environments where each agent autonomously decides its reasoning effort, exploration trajectories, query counts, and stopping criteria without artificial step caps. Detailed execution environments, model identifiers, and harness specifications are provided in Appendix~\ref{app:agent-configurations}.

\begin{table}[t]
\caption{Agent scores and rankings on Rank, Activation, Steering, and Overall across 20 tasks.}
\label{tab:agent-scores-ranked}
\centering\small\setlength{\tabcolsep}{4pt}\renewcommand{\arraystretch}{1.18}
\begin{tabular}{@{}lrrrrrrrr@{}}
\toprule
& \multicolumn{2}{c}{\cellcolor{overallwash}\textbf{Overall}} & \multicolumn{2}{c}{\cellcolor{rankwash}\textbf{Rank}} & \multicolumn{2}{c}{\cellcolor{activationwash}\textbf{Activation}} & \multicolumn{2}{c}{\cellcolor{steeringwash}\textbf{Steering}} \\
\cmidrule(lr){2-3}\cmidrule(lr){4-5}\cmidrule(lr){6-7}\cmidrule(l){8-9}
Agent & Score & Pos. & Score & Pos. & Score & Pos. & Score & Pos. \\
\midrule
\raisebox{-.12em}{\includegraphics[width=.9em,height=.9em,keepaspectratio]{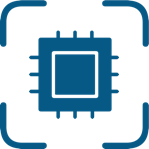}}\hspace{.35em}Expert~\citep{lin2023neuronpedia} & 85.56 & \textcolor{gray}{--} & 100.00 & \textcolor{gray}{--} & 98.92 & \textcolor{gray}{--} & 57.75 & \textcolor{gray}{--} \\
\midrule
\agentlogo{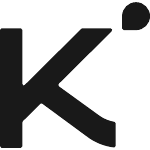}Kimi K3 & \textbf{65.82}\,\textcolor{gray}{\scriptsize$\pm$1.87} & \textcolor{gray}{1} & 74.30\,\textcolor{gray}{\scriptsize$\pm$5.60} & \textcolor{gray}{2} & \textbf{92.91}\,\textcolor{gray}{\scriptsize$\pm$1.07} & \textcolor{gray}{1} & 30.26\,\textcolor{gray}{\scriptsize$\pm$2.02} & \textcolor{gray}{2} \\
\agentlogo{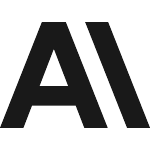}Claude Opus 5 & 65.41\,\textcolor{gray}{\scriptsize$\pm$0.41} & \textcolor{gray}{2} & \textbf{75.35}\,\textcolor{gray}{\scriptsize$\pm$0.14} & \textcolor{gray}{1} & 91.89\,\textcolor{gray}{\scriptsize$\pm$0.86} & \textcolor{gray}{3} & 28.99\,\textcolor{gray}{\scriptsize$\pm$2.01} & \textcolor{gray}{5} \\
\agentlogo{anthropic}Claude Sonnet 5 & 65.04\,\textcolor{gray}{\scriptsize$\pm$2.20} & \textcolor{gray}{3} & 73.45\,\textcolor{gray}{\scriptsize$\pm$2.81} & \textcolor{gray}{3} & 92.01\,\textcolor{gray}{\scriptsize$\pm$3.47} & \textcolor{gray}{2} & 29.66\,\textcolor{gray}{\scriptsize$\pm$2.68} & \textcolor{gray}{4} \\
\agentlogo{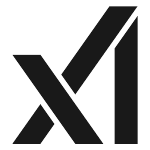}Grok 4.6 & 62.93\,\textcolor{gray}{\scriptsize$\pm$1.01} & \textcolor{gray}{4} & 67.21\,\textcolor{gray}{\scriptsize$\pm$3.07} & \textcolor{gray}{6} & 90.11\,\textcolor{gray}{\scriptsize$\pm$0.83} & \textcolor{gray}{5} & \textbf{31.47}\,\textcolor{gray}{\scriptsize$\pm$3.39} & \textcolor{gray}{1} \\
\agentlogo{anthropic}Claude Opus 4.8 & 62.57\,\textcolor{gray}{\scriptsize$\pm$5.42} & \textcolor{gray}{5} & 70.25\,\textcolor{gray}{\scriptsize$\pm$6.38} & \textcolor{gray}{4} & 90.90\,\textcolor{gray}{\scriptsize$\pm$3.56} & \textcolor{gray}{4} & 26.55\,\textcolor{gray}{\scriptsize$\pm$7.13} & \textcolor{gray}{6} \\
\agentlogo{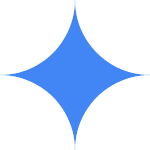}Gemini 3.8 Flash & 59.57\,\textcolor{gray}{\scriptsize$\pm$0.61} & \textcolor{gray}{6} & 69.49\,\textcolor{gray}{\scriptsize$\pm$5.07} & \textcolor{gray}{5} & 86.78\,\textcolor{gray}{\scriptsize$\pm$2.61} & \textcolor{gray}{7} & 22.43\,\textcolor{gray}{\scriptsize$\pm$1.39} & \textcolor{gray}{9} \\
\agentlogo{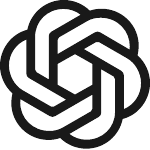}GPT-5.6 Sol & 57.28\,\textcolor{gray}{\scriptsize$\pm$0.86} & \textcolor{gray}{7} & 51.65\,\textcolor{gray}{\scriptsize$\pm$2.15} & \textcolor{gray}{8} & 90.03\,\textcolor{gray}{\scriptsize$\pm$0.91} & \textcolor{gray}{6} & 30.16\,\textcolor{gray}{\scriptsize$\pm$3.37} & \textcolor{gray}{3} \\
\agentlogo{openai}GPT-5.5 & 54.34\,\textcolor{gray}{\scriptsize$\pm$2.80} & \textcolor{gray}{8} & 55.63\,\textcolor{gray}{\scriptsize$\pm$5.28} & \textcolor{gray}{7} & 84.19\,\textcolor{gray}{\scriptsize$\pm$0.58} & \textcolor{gray}{9} & 23.19\,\textcolor{gray}{\scriptsize$\pm$3.50} & \textcolor{gray}{8} \\
\agentlogo{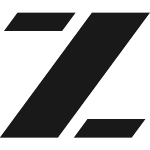}GLM-5.2 & 52.27\,\textcolor{gray}{\scriptsize$\pm$2.62} & \textcolor{gray}{9} & 47.66\,\textcolor{gray}{\scriptsize$\pm$5.75} & \textcolor{gray}{9} & 84.85\,\textcolor{gray}{\scriptsize$\pm$3.69} & \textcolor{gray}{8} & 24.28\,\textcolor{gray}{\scriptsize$\pm$1.69} & \textcolor{gray}{7} \\
\agentlogo{openai}GPT-5.6 Luna & 50.27\,\textcolor{gray}{\scriptsize$\pm$7.40} & \textcolor{gray}{10} & 46.92\,\textcolor{gray}{\scriptsize$\pm$14.91} & \textcolor{gray}{10} & 83.36\,\textcolor{gray}{\scriptsize$\pm$1.72} & \textcolor{gray}{10} & 20.54\,\textcolor{gray}{\scriptsize$\pm$5.76} & \textcolor{gray}{10} \\
\bottomrule
\end{tabular}
\par\smallskip{\footnotesize Scores are means over three runs. The $\pm$ values give the standard deviation across runs.}
\end{table}

\subsection{Evaluation Pipeline and Aggregation}
For each task, an agent conducts an autonomous, multi-turn investigation using the probe interface. Once the agent submits its chosen feature ID, the evaluation pipeline executes post-submission validation: measuring text-level activations on the frozen evaluation set and performing causal steering via greedy generation. Intervention scale $\alpha$ is calibrated against non-degeneration criteria on held-out prompts (Appendix~\ref{app:steering-protocol}), and steering outputs are rated in two passes by an automated GPT-4o judge using anonymized output triples (baseline, steered, and random control; Appendix~\ref{app:rubric}).

To evaluate consistency, every model--harness configuration conducts three independent, end-to-end investigations per task. We report the mean score and sample standard deviation ($\pm$) across the three runs, with benchmark-level scores averaging all 20 tasks equally. Across these experiments, we examine how effectively agents navigate the dictionary space to discover features, how they design contrastive probes and interpret empirical feedback, and how their selected features causally alter downstream generation.

\section{Results}

\subsection{Overall Performance}

Table~\ref{tab:agent-scores-ranked} presents the main benchmark results across the ten evaluated agent configurations, alongside the Neuronpedia Expert baseline. Overall performance reveals two overarching patterns. First, frontier agents exhibit substantial scientific discovery capabilities across the 20 tasks, with Kimi K3 achieving the highest composite Overall score of 65.82, closely followed by Claude Opus 5 (65.41) and Claude Sonnet 5 (65.04). Notably, different model families excel along distinct evaluative axes: Claude Opus 5 achieves the strongest dictionary-wide Activation Rank (75.35), Kimi K3 dominates Activation Selectivity (92.91), and Grok 4.6 leads in Causal Steering efficacy (31.47).

Second, while agents approach expert-level performance in distinguishing target concepts from contrastive texts, a persistent capability gap remains in causal intervention. On Activation Selectivity, top agents reach 92.91 compared to the Expert baseline of 98.92, indicating that agents can reliably identify features that selectively fire on target concepts. In sharp contrast, causal generation steering proves far more challenging: the top agent steering score reaches only 31.47 against Expert's 57.75. This divergence underscores that finding a correlated, selective feature does not readily translate into finding a causally potent steering vector. For instance, although GPT-5.6 Sol ranks third on Steering with a score of 30.16 and closely trails the category leader by only 1.31 points, its lower Activation Rank of 51.65 primarily accounts for its gap behind the top performers. 

Across repeated independent runs, agents exhibit stable overall rankings with modest variance, typically within 1--3 points in standard deviation, demonstrating that discovery capabilities remain consistent despite stochastic search trajectories (Appendix~\ref{app:aggregate}). Finally, comparing overall scores with general capability rankings on the Artificial Analysis Intelligence Index v4.2 reveals a strong rank correlation ($\rho=0.800$) across matching frontier models \citep{aa2026intelligence}, aligning well with general model capabilities while highlighting white-box interpretability as a distinct evaluative frontier (detailed in Appendix~\ref{app:aa-correlation}).

\takeaway{Frontier agents exhibit clear capability trade-offs in mechanistic discovery, but lag in causal validation.}{Across 20 single-feature discovery tasks, frontier models achieve strong activation selectivity but lag the curated expert baseline in causal steering (Overall 65.82 vs.\ 85.56, Steering 31.47 vs.\ 57.75). While agents can formulate informative contrastive probes to separate concepts in representation space, their current difficulty in identifying causally potent steering vectors highlights representational auditing as a critical bottleneck for future white-box RSI workflows.}

\subsection{How Agents Search and Test Candidates}
\label{sec:behavior}
\begin{table}[!htbp]
\caption{\textbf{Authored contrastive probes and case-level discovery scores on Portuguese (Layer 9).} For each agent, final benchmark scores on the submitted feature are shown alongside the contrastive probes designed during discovery and the resulting feature activations, illustrating candidate discrimination quality.}
\label{tab:agent-probe-controls}
\centering\small\setlength{\tabcolsep}{6pt}\renewcommand{\arraystretch}{0.98}
\begin{tabularx}{\linewidth}{@{}L{0.26\linewidth}>{\RaggedRight\arraybackslash}Xr@{}}
\toprule
\rowcolor{tablewash}
\textbf{Agent \& Case Scores} & \textbf{Agent-authored Probe Texts} & \textbf{Activation} \\
\midrule
\multicolumn{3}{@{}l@{}}{\raisebox{-.12em}{\includegraphics[width=.85em,height=.85em]{figures/logos/neuronpedia.png}}\hspace{.35em}\textbf{Neuronpedia Expert} \hfill \footnotesize\textnormal{Reference: \textbf{Rank} 100.0 $\,\mid\,$ \textbf{Act} 100.0 $\,\mid\,$ \textbf{Steer} 85.0 $\,\mid\,$ \textbf{Overall 95.0}}} \\
\midrule
\agentlogo{openai}\textbf{GPT-5.6 Sol} & \textbf{Portuguese:} Nas manhãs de domingo, costumo abrir\ldots & 7.04 \\
{\scriptsize Rank: 3.97 \quad Act: 100.0} & \textbf{English about Portuguese:} Portuguese is a Romance\ldots & 0.00 \\
{\scriptsize Steer: 88.8 \enspace \textbf{Ovr: 64.2}} & \textbf{Spanish translation:} Los domingos por la mañana, suelo abrir\ldots & 0.00 \\
\midrule
\agentlogo{anthropic}\textbf{Claude Opus 4.8} & \textbf{Portuguese:} O meu vizinho comprou um carro novo\ldots & 11.94 \\
{\scriptsize Rank: 30.8 \quad Act: 91.7} & \textbf{French about Brazil:} Le Brésil est le plus grand pays\ldots & 0.00 \\
{\scriptsize Steer: 84.4 \enspace \textbf{Ovr: 68.9}} & \textbf{Spanish translation:} Mi vecino compró un coche nuevo\ldots & 0.00 \\
\midrule
\agentlogo{anthropic}\textbf{Claude Opus 5} & \textbf{Portuguese:} Hoje de manhã fui à padaria comprar\ldots & 3.81 \\
{\scriptsize Rank: 0.08 \quad Act: 41.7} & \textbf{Spanish translation:} Esta mañana fui a la panadería\ldots & 1.19 \\
{\scriptsize Steer: 0.00 \enspace \textbf{Ovr: 13.9}} & \textbf{English translation:} This morning I went to the bakery\ldots & 4.50 \\
\midrule
\agentlogo{kimi}\textbf{Kimi K3} & \textbf{Portuguese:} Ontem fui ao mercado comprar frutas\ldots & 26.33 \\
{\scriptsize Rank: 100.0 \enspace Act: 100.0} & \textbf{English translation:} Yesterday I went to the market\ldots & 0.00 \\
{\scriptsize Steer: 85.0 \enspace \textbf{Ovr: 95.0}} & \textbf{Spanish translation:} Ayer fui al mercado a comprar\ldots & 1.80 \\
 & \textbf{English about Portugal:} Portugal is a country\ldots & 0.00 \\
\bottomrule
\end{tabularx}
\end{table}

How do agents formulate hypotheses, construct contrastive probes, and navigate empirical feedback? Table~\ref{tab:agent-probe-controls} illustrates this process across four independent Portuguese investigations in the layer-9 SAE, linking the contrastive probes designed during discovery to final benchmark scores.

\textbf{Cross-lingual probing and evidence interpretation.} The probes reveal clear differences in concept discrimination. Sol, Opus 4.8, and Kimi K3 successfully identify features that distinguish Portuguese from related languages (Spanish) and non-target language descriptions (English or French), with zero or negligible activation on controls. However, their final Rank scores reflect an important trade-off between selectivity and activation strength: Sol's feature achieves clean separation with an Activation score of 100.00 and strong causal steering of 88.75, but its target activation is modest at 7.04, placing it lower in the dictionary-wide ranking with a Rank score of 3.97. In contrast, Kimi K3 identifies a prominent direction with much stronger target activation of 26.33, attaining both perfect separation and a leading Rank score of 100.00. Meanwhile, Opus 5 illustrates an evidence-interpretation failure: despite authoring matched Portuguese, Spanish, and English probes, its selected feature activates more strongly on the English control at 4.50 than on the Portuguese target at 3.81, leading to an overall score collapse down to 13.91. Detailed probe suites and complete search traces appear in Appendices~\ref{app:full-authored-probes} and~\ref{app:extended-investigations}.

\textbf{Separating concepts from format and substring distractors.} Beyond cross-lingual contrasts, agents construct targeted counterexamples to rule out superficial string matching and format-level spurious correlations. In the \textit{Cat} discovery task, Claude Opus 5 encounters a candidate feature that activates strongly on genuine feline sentences at 32.92, but also fires aggressively on unrelated compound words such as \emph{Copycat killer} at 44.00 and \emph{Catalytic converter} at 24.54. By explicitly designing these morphological negative controls, Opus 5 identifies that the candidate is merely detecting the substring ``cat'' rather than the animal concept, and successfully rejects it in favor of the monosemantic Expert feature, as detailed in Appendix Table~\ref{tab:trace-cases}. Conversely, GLM-5.2 demonstrates an evidence-misinterpretation failure on the \textit{Clinical symptom} task: while it designs diagnostic controls to rule out general medical bureaucracy, its final candidate still fires at 70.25 on a patient medical history entirely devoid of symptoms, as detailed in Appendix Table~\ref{tab:clinical-contrast-probes}. GLM-5.2 erroneously downplays this strong non-symptom response as negligible in its report, mistakenly selecting a feature driven by clinical document formatting rather than actual symptoms.

\textbf{Feature purity versus generation degeneration.} Agents also face fundamental trade-offs between concept selectivity and downstream behavioral impact. In the \textit{Real estate} task, agents diverge on how to balance candidate purity against coverage, illustrated in Appendix Table~\ref{tab:real-estate-agent-comparison}. Sol and Grok 4.6 deliberately reject a broad feature candidate due to minor leakages on non-housing ads, selecting a strictly selective feature instead. In contrast, Kimi K3 prioritizes broad listing coverage and accepts this leakier candidate. Consequently, Kimi's chosen feature achieves higher target relevance and causal steering, but at the cost of substantial generation degeneration, rising from 32.5\% to 52.5\%. These diverse strategies illustrate that autonomous discovery requires not only finding selective features, but also arbitrating trade-offs between precision, coverage, and output stability.

\begin{figure}[!ht]
\centering\includegraphics[width=\linewidth]{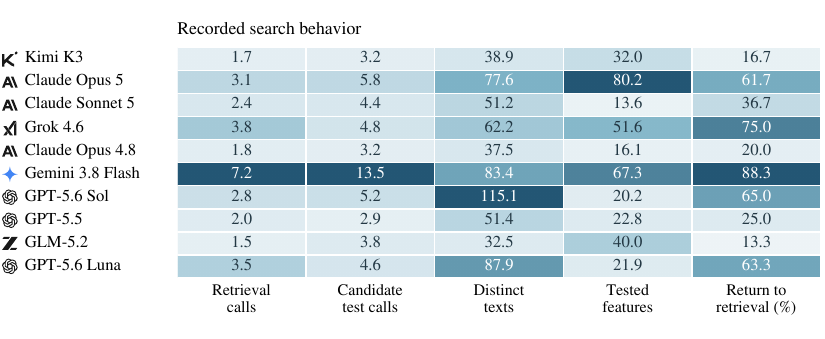}
\caption{\textbf{Autonomous scientific search workflows across frontier agents.} Metrics report average behavior per investigation across 20 tasks: number of authored probe requests, unique authored probe texts, directly tested feature candidates, and the percentage of episodes where the agent resumed candidate retrieval after direct testing. Column color scales are normalized independently.}
\label{fig:search-behavior-all}
\end{figure}

Behind broad summary metrics, agents exhibit fundamentally distinct scientific investigation philosophies, as reflected in their search workflows in Figure~\ref{fig:search-behavior-all}. Rather than following a uniform trial-and-error strategy, models diverge along two critical dimensions: candidate breadth versus probe depth, and deliberate verification versus heuristic selection.

\textbf{Exploration breadth vs.\ probe depth.} Sol prioritizes textual diversity over dictionary span, authoring extensive probe variations averaging 115.1 texts across sentence lengths, topical contexts, and cross-lingual translations, while testing a relatively small pool of 20.2 candidate features. Conversely, Claude Opus 5 behaves as a broad screener, comparing over four times as many candidate features at 80.2 across fewer probe texts averaging 77.6. Meanwhile, Kimi K3 adopts a highly targeted strategy: making far fewer retrieval queries and rarely returning to retrieval once a promising candidate is identified, yet achieving top overall discovery scores through precise hypothesis formulation.

\textbf{Active validation vs.\ heuristic selection.} Even when agents converge on identical features, their underlying scientific rigor differs sharply. For example, while Claude Opus 5 and Sonnet 5 select the canonical Expert feature at nearly identical rates, their decision traces reveal opposite behaviors: Opus 5 acts as an active hypothesis-tester, repeatedly authoring dedicated probes to directly evaluate the Expert candidate before either selecting or deliberately rejecting it in favor of an alternative. In contrast, Sonnet 5 frequently encounters the Expert candidate during initial retrieval but overlooks it without direct validation, relying instead on initial ranking heuristics. Because Expert identity is fully anonymized during discovery, these patterns underscore that autonomous agents vary substantially in scientific thoroughness, spanning from active experimental verification to passive reliance on initial retrieval outputs, with comprehensive search outcomes detailed in Appendix~\ref{app:supplementary-search}.

\takeaway{Search strategy matters, but rigorous hypothesis testing determines discovery success.}{Agents explore differently: Sol varies texts, Opus screens candidates, and Kimi targets precise hypotheses. Yet finding effective features depends less on search volume than on designing informative counterexamples, avoiding surface distractors, and accurately reading experimental results.}

\subsection{What Activation Scores Reveal}

\begin{figure}[!t]
    \centering\includegraphics[width=\linewidth]{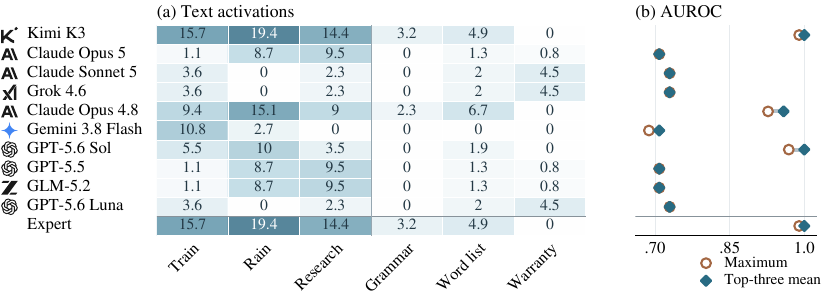}
    \caption{\textbf{Generalization and aggregation sensitivity of discovered Portuguese features.} (a)~Activations across representative positive texts (Train, Rain, Research) and contrastive controls (Grammar, Word list, Warranty), demonstrating that identical AUROC separation (e.g., Sol vs.\ Expert) conceals major differences in absolute activation strength and dictionary rank. (b)~AUROC comparison between maximum token activation and top-three-token mean aggregation; token averaging dampens isolated lexical spikes in control texts, reliably isolating continuous semantic activation.}
    \label{fig:activation-distributions}
    \end{figure}

While agent-authored probes guide candidate selection during search, our shared benchmark suite objectively evaluates how the submitted features generalize across unseen contexts. Following the Portuguese investigations examined in Table~\ref{tab:agent-probe-controls}, Figure~\ref{fig:activation-distributions} tests the identical submitted candidate features against the standardized Portuguese evaluation suite of eight held-out positive sentences and twelve contrastive controls (full texts in Appendix Table~\ref{tab:matched-activation}). This controlled blind evaluation exposes substantial disparities in generalization and selectivity that remain invisible during internal search.

\textbf{Coverage gaps and spurious activations.} Features that appeared promising under agent probes often falter on held-out benchmark texts. As shown in Figure~\ref{fig:activation-distributions}a, candidates selected by Sonnet, Grok, and Luna remain completely inactive on natural target sentences such as the rain passage, yet exhibit spurious activations on irrelevant English controls such as warranty statements. Similarly, Gemini's submitted feature fails to fire on the research passage and misses multiple positives across the broader test set. These coverage gaps demonstrate that surface-level probe validation frequently under-specifies concept boundaries, allowing models to select features tied to narrow lexical cues rather than the intended global semantic concept.

\textbf{High selectivity vs.\ weak activation strength.} A high scalar Activation score can mask critical differences in feature strength. Both Sol and Expert achieve a perfect AUROC of 1.000 across the full evaluation suite, cleanly ranking every positive text above every control text. However, their absolute activation magnitudes tell a fundamentally different story. Sol's feature is exceptionally clean, remaining completely inactive at 0.00 across almost all control texts, but its target activation on Portuguese sentences is very modest, averaging only 3.5--10.0 compared to 14.4--19.4 for Expert and 26.3 for Kimi K3 in Table~\ref{tab:agent-probe-controls}. Because its target signal is relatively weak, the feature fails to stand out against competing features across the full dictionary, collapsing Sol's Activation Rank score to only 3.97 versus 100.00 for Expert. This contrast illustrates that a feature can exhibit near-perfect selectivity while remaining too weak to serve as a prominent, primary representation of the target concept, explaining why rank and selectivity must be evaluated together.

\textbf{Robustness via top-three-token aggregation.} Figure~\ref{fig:activation-distributions}b highlights why sequence-level aggregation is critical when evaluating concept selectivity. Relying solely on the maximum token activation leaves metrics vulnerable to isolated token outliers. For example, in Sol's vocabulary control, an isolated Portuguese loanword produces a sharp single-token spike of 5.56, whereas the full target sentence maintains a moderate per-token response of 2.96. Under maximum aggregation, this spurious spike degrades Sol's AUROC to 0.969. In contrast, aggregating the top-three token activations amortizes transient spikes, raising Sol's AUROC to 1.000 and Opus 4.8's from 0.927 to 0.958 (detailed in Appendix Table~\ref{tab:activation-peak}). This verifies that multi-token averaging effectively filters isolated morphological artifacts while rewarding coherent semantic activation across the text. Complete activation matrices across all tasks appear in Appendix Figure~\ref{fig:activation-controls-all}.

\subsection{Steering and Instruction Preservation}
\label{sec:steering-text}
\begin{figure}[!htbp]
    \centering\includegraphics[width=\linewidth]{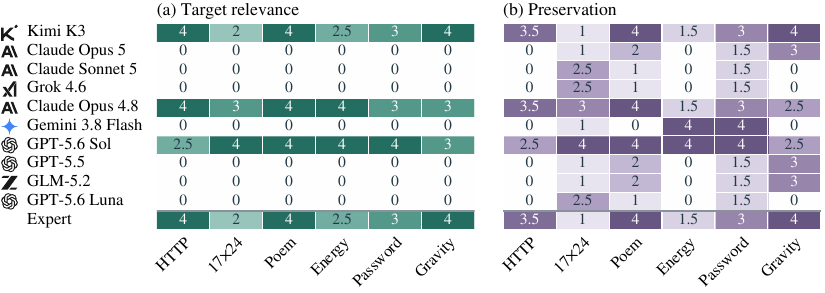}
    \caption{\textbf{Causal steering and instruction preservation under Portuguese feature intervention.} Evaluations across six diverse instructions: HTTP request explanation, 17$\times$24 arithmetic, 4-line poem, clean energy essay, password generation, and gravity definition. (a)~\textit{Target relevance} (0--4 scale) measures the intensity of induced Portuguese expression; (b)~\textit{Instruction preservation} (0--4 scale) measures fulfillment of the original task constraints without degeneration or incoherence.}
    \label{fig:matched-steering}
\end{figure}

While activation scores measure representational correlation, activation steering intervenes directly on inference to evaluate causal control. Following the identical Portuguese features examined in Table~\ref{tab:agent-probe-controls} and Figure~\ref{fig:activation-distributions}, Figure~\ref{fig:matched-steering} evaluates how intervening along each agent's submitted direction ($h \leftarrow h + \alpha d_f$) affects base model generation across six evaluation instructions. Each generation is rated on a 0--4 scale along two orthogonal axes: \textit{target relevance} (degree of Portuguese expression induced) and \textit{instruction preservation} (retention of user constraints such as length, topic, and formatting).

Figure~\ref{fig:matched-steering} reveals a sharp divide in causal efficacy across submitted features. While many features achieved strong activation separation in Figure~\ref{fig:activation-distributions}, only a small fraction reliably steer generation. Features submitted by Sol, Opus 4.8, and Kimi K3, which discovered the Expert feature 41424, exhibit high target relevance across instructions, successfully flipping model outputs into fluent Portuguese, as illustrated in Table~\ref{tab:steering-short}. In contrast, features selected by Opus 5, Sonnet 5, Grok 4.6, and Luna yield flat zero target relevance across all six prompts, indicating that their correlated activations fail to exert causal influence on downstream generation.

\begin{table}[!t]
\caption{Opening excerpts under Portuguese steering.}
\label{tab:steering-short}
\centering\small\setlength{\tabcolsep}{5pt}\renewcommand{\arraystretch}{1.05}
\begin{tabular}{@{}ll@{}}
\toprule
\multicolumn{2}{@{}l@{}}{\textbf{Prompt:} Introduce yourself in two sentences.} \\
\midrule
Selected by & Opening excerpt \\
\midrule
No steering & \textcolor{casegrayink}{I am Gemma, an open-weights AI assistant\ldots} \\
\raisebox{-.12em}{\includegraphics[width=.85em,height=.85em]{figures/logos/neuronpedia.png}}\hspace{.25em}Expert / \agentlogo{kimi}Kimi K3 & \textcolor{caseteal}{Olá! Eu sou o Gemma, um modelo de linguagem\ldots} \\
\agentlogo{openai}GPT-5.6 Sol & \textcolor{caseteal}{Olá! Eu sou Gemma, um modelo de linguagem\ldots} \\
\agentlogo{anthropic}Claude Opus 4.8 & \textcolor{caseteal}{Olá! Eu sou um modelo de linguagem grande\ldots} \\
\agentlogo{openai}GPT-5.6 Luna & \textcolor{casegrayink}{Hello! I am Gemma, an open-weights AI assistant\ldots} \\
\bottomrule
\end{tabular}

\end{table}

Furthermore, generation steering reveals subtle trade-offs between target induction and output quality. While Steering isolates the net gain in target expressions, a high score does not guarantee well-formed completions if user constraints are violated. Across the full 20-instruction benchmark suite, Sol achieves a leading Steering score of 88.75 with high preservation of 3.525, whereas Opus 4.8 attains 84.38 with lower preservation of 3.050 due to repetition and length drift. For instance, on a requested four-line rhyming poem, Sol outputs four coherent Portuguese lines, while Opus 4.8 generates eleven lines with repetitive loops, and Expert generates six. In more extreme failures like the clinical symptom task, GLM-5.2's intervention completely disrupts instruction following, converting an innocent self-introduction into a rambling patient medical history. Complete generated responses, preservation ratings, and degeneration checks appear in Appendix~\ref{app:agent-output-atlas}.

\takeaway{Causal steering separates superficial activation correlates from actionable mechanisms.}{While multiple agents discover features that achieve near-perfect activation separation on static texts, very few translate into reliable steering directions. Effective causal intervention requires precise semantic modulation without triggering degeneration, instruction drift, or formatting collapse, marking causal steering as the most rigorous test of autonomous interpretability research.}

\section{Related Work}

\paragraph{Mechanistic interpretability: From neurons to SAE features.}
Mechanistic interpretability studies model representations through neuron analysis, circuit analysis, causal tracing, and knowledge attribution \citep{elhage2021mathematical,meng2022locating,wang2022interpretability,geva2021transformer,dai2022knowledge,song2024does,yu2024neuron}. However, polysemanticity, where individual neurons activate across unrelated concepts, fundamentally limits the interpretability of raw hidden states \citep{bricken2023monosemanticity}. Sparse Autoencoders (SAEs) address this barrier by decomposing internal activations into sparse combinations of learned feature directions \citep{cunningham2023sparse,gao2024scaling,templeton2024monosemanticity}. These features support causal intervention \citep{marks2024circuits}, with steering effectiveness depending on their influence on model outputs \citep{arad-etal-2025-saes,shu-etal-2025-beyond,wang2026utility}. Recent selection methods use correlations with task performance, supervision, or similarities between features to improve steering \citep{cho2026corrsteer,jorgensen2026steering,liu2026nifs}. Pretrained libraries such as Gemma Scope make large dictionaries accessible \citep{lieberum2024gemmascope}, while feature splitting, absorption, and composition complicate the mapping between features and concepts \citep{chanin2024absorption,leask2025canonical}.

\paragraph{Automated interpretation and SAE benchmarking.}
Automated methods use language models to explain neurons and SAE features from activating examples \citep{bills2023language,paulo2025millions}. Agents extend this process through interactive experiments \citep{shaham2024maia,han2026sage,bissell2025scribe}, feature discovery \citep{marinllobet2026automated}, and circuit analysis \citep{khan2026circuitexplainers}. Other agents investigate hidden model behaviors using auditing tools \citep{bricken2025auditing,sheshadri2026auditbench}. Agents also assess mechanistic interpretability research by examining papers, code, and data for experimental coherence, reproducibility, and generalizability \citep{bai2026story}. Existing evaluations measure SAE quality and feature recovery \citep{karvonen2025saebench,venhoff2024sage}, interpretability without generated explanations \citep{paulo2025withoutexplanations}, concept disentanglement \citep{huang2024ravel}, circuit identification \citep{mueller2025mib}, and concept detection and steering \citep{wu2025axbench}. Studies also examine whether SAE metrics reliably distinguish feature quality and learned structure \citep{chanin2026reliable,heap2026metrics}. \bench{} evaluates agents' ability to search pretrained SAE dictionaries using contrastive probes, with scores for activation rank, activation selectivity, and steering.

\paragraph{Agents for AI research and development.}
Recursive self-improvement motivates research on agents that can improve their own software and contribute to AI development \citep{zhang2025darwin,mahmoud2026rsibench}. Existing benchmarks assess software and machine learning engineering \citep{tan2026swetouch,chan2024mlebench}, paper reproduction \citep{starace2025paperbench}, scientific insight rediscovery \citep{wang2026fire}, research proposal assessment \citep{ho2026soundnessbench}, and literature discovery \citep{xiong2026autoresearchbench}. Other evaluations cover scientific data analysis \citep{chen2025scienceagentbench}, data curation and post-training \citep{meng2026rsibenchdata,rank2026posttrainbench,tan2026posttrainbench0}, and broader experimental research workflows \citep{edelman2026airs,wang2026researchclawbench}. Together, these studies assess distinct skills involved in research and development. \bench{} extends this evaluation to experiments on trained model representations, measuring how agents design probes, select SAE features, and obtain features that influence generation.

\section{Limitations}
\label{sec:limitations}

While \bench{} establishes a standardized foundation for evaluating autonomous mechanistic interpretability agents, several limitations offer directions for future work. First, our benchmark focuses on concept-driven single-feature discovery within Gemma-2-9B-IT across two layers of Gemma Scope residual-stream SAEs under a fixed probe interface. Expanding evaluations to diverse model families, broader dictionary widths, multi-feature circuit discovery, and open-ended hypothesis generation without pre-specified concepts will provide a more comprehensive view of agent capabilities. Second, candidate evaluation relies on fixed text suites, frozen expert baselines, and automated LLM-as-a-judge assessments for steering; incorporating human-in-the-loop validation, multiple reference features, and multi-judge ensembles can further minimize potential rating noise. Finally, while \bench{} benchmarks post-hoc feature auditing, integrating agent-discovered features directly into downstream model editing, unlearning, or continual alignment loops represents an essential next step toward realizing fully closed-loop recursive self-improvement.

\section{Conclusion}

\bench{} establishes a standardized benchmark to evaluate AI agents as scientists conducting autonomous mechanistic interpretability research. Across ten frontier agent configurations and 20 tasks, our evaluations reveal that while agents approach the expert reference baseline in distinguishing concepts in activation space, a substantial capability gap persists in causal steering. Behavioral analyses demonstrate that effective discovery hinges not on exploration volume, but on rigorous hypothesis testing, designing informative counterexamples, and accurately interpreting experimental feedback. By bridging mechanistic interpretability with autonomous AI research, \bench{} provides an essential, auditable foundation toward verifying internal representations in recursive self-improvement loops.

\section*{Reproducibility Statement}
Appendix~\ref{app:setup} specifies the tasks, agents, and evaluation protocol. Appendix~\ref{app:search} follows the investigations. Appendices~\ref{app:activation} and~\ref{app:steering} provide activation measurements and complete generations. Appendix~\ref{app:aggregate} compares tasks and repeated runs. The underlying records preserve model identifiers and execution traces.

\section*{AI Use Statement}
Generative AI tools were used solely for writing assistance, text polishing, and editorial phrasing throughout the preparation of this manuscript. All research conceptualization, benchmark design, experimental execution, data collection, and analytical interpretations were conducted entirely by the human authors.

\begingroup
\expandafter\def\expandafter\UrlBreaks\expandafter{\UrlBreaks\do\-}
\setlength{\bibsep}{2pt plus 1pt minus 1pt}
\bibliography{references}
\bibliographystyle{template/iclr2027_conference}
\endgroup

\clearpage
\appendix
\raggedbottom
\makeatletter
\setlength{\@fptop}{0pt}
\setlength{\@fpsep}{20pt}
\setlength{\@fpbot}{0pt plus 1fil}
\makeatother
\renewcommand{\tabularxcolumn}[1]{L{#1}}
\renewcommand{\arraystretch}{1.13}
\section{Tasks and Experimental Protocol}
\label{app:setup}
The appendix follows the experiment from task construction to agent investigation, activation testing, and generated outputs. Task and feature identifiers are retained in measurement tables to make the records traceable. The narrative describes candidates by their observed responses.
\subsection{Task Inventory}
\label{app:task-inventory}
The benchmark uses 20 concept--layer tasks and 17 concepts. Earnings reports, portfolio allocation, and tax filing each occur at both layers. An Expert ID is local to its stated SAE checkpoint. Crucially, all 20 Expert features serve as frozen reference baselines: where public steering presets are documented on Neuronpedia \citep{lin2023neuronpedia} alongside Gemma Scope \citep{lieberum2024gemmascope} (e.g., Cat), they are adopted directly; for remaining concepts, features are validated via standard expert curation workflows as detailed below.
\begin{table}[H]
\caption{Task concepts and Expert IDs. Each SAE contains 131,072 features.}
\label{tab:task-landscape}
\centering\small
\begin{tabular}{rlr}
\toprule
Layer & Concept & Expert ID \\
\midrule
20 & archaeological excavation & 7256 \\
9 & cat & 62610 \\
20 & clinical symptom reports & 3927 \\
9 & earnings reports & 131024 \\
20 & earnings reports & 100747 \\
9 & French & 105738 \\
9 & gardening advice & 93406 \\
9 & German & 33987 \\
9 & job postings & 91086 \\
9 & Latin & 7659 \\
20 & pharmaceutical dosing & 104583 \\
9 & portfolio allocation & 77390 \\
20 & portfolio allocation & 101617 \\
9 & Portuguese & 41424 \\
20 & real estate listings & 27182 \\
9 & Spanish & 94329 \\
9 & tax filing language & 18713 \\
20 & tax filing language & 78694 \\
9 & Turkish & 99383 \\
20 & weather forecasts & 44494 \\
\bottomrule
\end{tabular}

\end{table}

\subsection{Expert Construction and Evaluation Reuse}
\label{app:reference-construction}
The Expert features serve as established reference baselines, directly adopted from public Neuronpedia presets where available (e.g., Cat originates from Neuronpedia's primary steering preset). For the remaining tasks, candidate features were validated using standard expert curation workflows: candidates were retrieved from six positive and six negative construction probes, then verified on eight positive, eight hard-negative, and four neutral texts to confirm concept fidelity. All Expert IDs occur in the retained validation records, and their activation case lists and matching metrics are fully traceable.

The evaluation texts are unavailable to agents but were used during Expert construction. Because Expert features represent canonical, pre-identified directions from official presets and verified workflows, they provide an objective, fixed reference point rather than an artificially tuned competitor. Gemma Scope supplies the SAE checkpoints, and Neuronpedia documents their features. The text suites and Expert selection procedure are specified here as benchmark components. The unified Expert steering evaluation regenerates answers using the task's frozen scale and the common 4o judge. Rank compares against Expert. Activation and Steering use each feature's measurements directly.

\subsection{Agent Configurations and Repeated Runs}
\label{app:agent-configurations}
\begin{table}[H]
\caption{Agent models and execution harnesses.}
\label{tab:agent-configurations}
\centering\small\setlength{\tabcolsep}{5pt}
\begin{tabularx}{\linewidth}{@{}l l >{\ttfamily}X@{}}
\toprule
\rowcolor{tablewash}Agent & Harness & \normalfont Model identifier \\
\midrule
\agentlogo{anthropic}Claude Opus 5 & Cursor & claude-opus-5-thinking-high \\
\agentlogo{anthropic}Claude Sonnet 5 & Cursor & claude-sonnet-5-thinking-high \\
\agentlogo{kimi}Kimi K3 & Cursor & kimi-k3-high \\
\agentlogo{anthropic}Claude Opus 4.8 & Cursor & claude-opus-4-8-thinking-high \\
\agentlogo{xai}Grok 4.6 & Cursor & cursor-grok-4.6-high \\
\agentlogo{gemini}Gemini 3.8 Flash & Cursor & gemini-3.8-flash-high \\
\agentlogo{openai}GPT-5.6 Sol & Codex & gpt-5.6-sol \\
\agentlogo{openai}GPT-5.5 & Cursor & gpt-5.5-high \\
\agentlogo{zai}GLM-5.2 & Cursor & glm-5.2-high \\
\agentlogo{openai}GPT-5.6 Luna & Codex & gpt-5.6-luna \\
\bottomrule
\end{tabularx}

\par\smallskip
\begin{minipage}{\linewidth}\footnotesize\RaggedRight
Each configuration has 60 scored episodes, covering three investigations of each task. Cursor model identifiers encode the high reasoning setting. Codex receives an explicit \texttt{reasoning\_effort=high} argument.
\end{minipage}
\end{table}

Each configuration contributes three complete runs over 20 tasks. Failed attempts (e.g., tool syntax errors) are replaced by retries, with exactly one scored result per episode. Agents interact with the environment using identical task specifications, probe interfaces, a dedicated writable workspace, and disabled network access. Cursor and Codex traces are audited to ensure full compliance. The evaluation runner applies a standard 60-minute execution timeout per task.

\subsection{Metric Definitions and Worked Example}
\label{app:metric-dictionary}
Benchmark scores implement Equations~\ref{eq:rank-current}--\ref{eq:overall-score}. Text activations pool the three largest non-special-token responses. AUROC credits ties at half a point against pooled hard-negative and neutral controls.
\begin{table}[H]
\caption{Component scores for GPT-5.6 Sol's Portuguese selection.}
\label{tab:worked-score}
\centering\small\setlength{\tabcolsep}{7pt}
\begin{tabular}{lrr}
\toprule
Measurement or score & \agentlogo{openai}GPT-5.6 Sol: 49607 & Expert: 41424 \\
\midrule
Mean positive rank & 642.625 & 13.000 \\
AUROC & 1.000 & 1.000 \\
Target Effect & 0.8875 & 0.8500 \\
\midrule
Rank & 3.97 & 100.00 \\
Activation & 100.00 & 100.00 \\
Steering & 88.75 & 85.00 \\
Overall & 64.24 & 95.00 \\
\bottomrule
\end{tabular}

\end{table}

The score export records all 600 episodes, including raw activation ranks, AUROC, and Target Effect, preserving each run and submitted feature ID. Exact ID recovery is reported as a separate diagnostic. Table~\ref{tab:worked-score} provides a worked step-by-step calculation illustrating how these metrics are evaluated for both Expert and agent submissions.

\subsection{Experiment Accounting}
\label{app:accounting}
In total, the benchmark archive contains 600 scored episodes, 4,873 probe calls, and 49,407 text entries across the evaluated agents. Steering evaluation covers 124 task--direction pairs and 10,940 generations across calibration and final evaluation, evaluated over two independent passes by GPT-4o (yielding 14,880 condition ratings).

\subsection{Steering Generation Protocol}
\label{app:steering-protocol}
\begin{table}[H]
\caption{Frozen generation and steering settings. Counts refer to one task--direction evaluation unless stated otherwise.}
\label{tab:steering-protocol}
\centering
\small
\arrayrulecolor{tablerule}
\begin{tabularx}{\linewidth}{>{\RaggedRight\arraybackslash}L{0.28\linewidth}X}
\toprule
\rowcolor{tablewash}
Setting & Value \\
\midrule
Generator & \texttt{google/gemma-2-9b-it} with greedy decoding \\
Intervention & Add $\alpha W_{\mathrm{dec}}[f,:]$ at every token position \\
Expert scale & Frozen task $\alpha_E$. Values across the 20 tasks: $\{45,60,120,160,240\}$ \\
Alternative scale grid & $\{0.5,0.75,1,1.25,1.5\}\alpha_E$. If none yields at least 90\% non-degenerate calibration outputs, try $\{0.0625,0.125,0.25,0.375\}\alpha_E$ \\
Calibration data & 5 prompts per attempted scale. Deterministic cue score only \\
Scale tie-break & Cue success rate, then mean cue score, then smaller $\alpha$ \\
Evaluation data & 20 instructions with baseline, feature, and matched-random generation for each \\
Generation length & 64 tokens for 15 tasks, 128 for 4 language tasks, and 192 for the cat task \\
Random control & Seed-0 Gaussian vector rescaled to the submitted decoder row's norm \\
Judge & 4o, two separately shuffled passes, temperature 0 \\
\bottomrule
\end{tabularx}
\end{table}

\paragraph{Scale selection and final scoring.} To enable fair causal comparison while preventing output degeneration, the evaluator calibrates the intervention strength $\alpha$ for each submitted feature using a principled grid search on held-out calibration prompts prior to benchmark scoring. Expert features retain their frozen task reference scale $\alpha_E$ (Table~\ref{tab:steering-protocol}), adopting Neuronpedia's public steering preset for Cat and established task calibration scales for remaining tasks. For agent-submitted alternative features, the pipeline searches over the layer-specific candidate grid in Table~\ref{tab:steering-protocol} (e.g., $\alpha \in \{80, 120, 160, 200, 240\}$ for Layer 9).

The calibration selects $\alpha$ via a three-stage filter on five held-out calibration prompts:
\begin{enumerate}
\setlength{\itemsep}{2pt}
\setlength{\parskip}{0pt}
\item \textbf{Non-degeneration filter:} The candidate scale must yield $\ge 90\%$ non-degenerate generations (all 5 calibration outputs must pass coherence and formatting checks without repetitive loops).
\item \textbf{Target cue optimization:} Among retained scales passing the degeneration check, the scale maximizing the target keyword cue success rate (and tie-broken by mean cue score) is selected.
\item \textbf{Minimal perturbation tie-break:} If multiple scales achieve identical target cue performance without degeneration, the smaller $\alpha$ is preferred to minimize unnecessary perturbation of the representation space.
\end{enumerate}
The selected scale is then frozen and applied across all 20 final evaluation instructions. The norm-matched random direction control uses the identical calibrated scale. Intervention scales for selected generation cases appear in Table~\ref{tab:case-intervention-scales}.

\subsection{Steering Evaluation Rubric and Judge Prompts}
\label{app:rubric}
An automated GPT-4o judge evaluates steered completions against the user instruction and target concept. To ensure blinded and impartial evaluation, the judge receives the generation instruction along with three candidate completions labeled A, B, and C in pseudo-randomized order, concealing whether an output originated from baseline inference, the submitted SAE feature intervention, or the norm-matched random-direction control.

For each completion, the judge independently assigns:
\begin{itemize}
\setlength{\itemsep}{2pt}
\setlength{\parskip}{0pt}
\item \textbf{Target Relevance (0--4):} Quantifies the degree to which the target concept is induced (0 = none, 1 = weak, 2 = implicit, 3 = explicit, 4 = dominant).
\item \textbf{Task Preservation (0--4):} Measures adherence to original constraints (e.g., formatting, length, answering user queries).
\item \textbf{Degeneration (Boolean):} Flags severe output collapse, such as repetitive token loops or incoherent formatting.
\end{itemize}
Ratings are averaged across the 20 evaluation prompts and two independent evaluation passes per prompt at temperature 0, with strict JSON output formatting. The complete system prompt, user prompt template, and concept-specific scoring criteria are detailed below.

\noindent\textbf{Judge messages.}
The templates below reproduce the scoring instructions. Angle-bracket fields are replaced by the concept criteria, generation prompt, and three answers. Line wrapping is adjusted for display. The user message requests one rating for each of A, B, and C.

\begin{judgeprompt}{promptblue}{promptbluewash}{System prompt}
You are evaluating activation steering for one SAE feature.\newline
The intended concept is \promptfield{concept}.

For each candidate output, independently score:

\textbf{\texttt{- target\_relevance (0-4):}}\newline
\hspace*{.8em}0 none;\newline
\hspace*{.8em}1 weak or ambiguous evidence;\newline
\hspace*{.8em}2 implicit but recognizable target content;\newline
\hspace*{.8em}3 explicit target content or multiple specific cues;\newline
\hspace*{.8em}4 dominant and repeated target content.

\textbf{\texttt{- task\_preservation (0-4):}}\newline
\hspace*{.8em}0 ignores the user's task;\newline
\hspace*{.8em}1 barely addresses it;\newline
\hspace*{.8em}2 partially answers it;\newline
\hspace*{.8em}3 mostly answers it;\newline
\hspace*{.8em}4 fully answers it.

\textbf{\texttt{- degenerate:}} true only for broken, severely repetitive, incoherent, or unusably truncated output.

\textbf{Strong target evidence:} \promptfield{strong\_evidence}\newline
\textbf{Do not count as sufficient evidence:} \promptfield{insufficient\_evidence}

\textbf{Return strict JSON only:}\newline
{\ttfamily\footnotesize
\{\newline
\hspace*{1em}"ratings": [\{\newline
\hspace*{2em}"label": "A", "target\_relevance": 0,\newline
\hspace*{2em}"task\_preservation": 0, "degenerate": false,\newline
\hspace*{2em}"reason": "short reason"\newline
\hspace*{1em}\}]\newline
\}}
\end{judgeprompt}

\begin{judgeprompt}{promptteal}{prompttealwash}{User prompt}
\textbf{User task:}\newline
\promptfield{evaluation\_prompt}

\textbf{Output A:} \promptfield{answer\_A}\newline
\textbf{Output B:} \promptfield{answer\_B}\newline
\textbf{Output C:} \promptfield{answer\_C}

Evaluate every output above exactly once. Return exactly three ratings with labels A, B, and C; do not omit, rename, or add labels.
\end{judgeprompt}

\medskip
\noindent\textbf{Concept criteria.}
Table~\ref{tab:judge-concept-criteria} gives the exact values inserted into the system message. The three concepts evaluated at both SAE layers use the same wording at each layer.

\begingroup
\fontsize{9}{10}\selectfont
\setlength{\tabcolsep}{5pt}
\renewcommand{\arraystretch}{1.0}
\begin{longtable}{@{}L{0.22\linewidth}L{0.36\linewidth}L{\dimexpr0.42\linewidth-20pt\relax}@{}}
\caption{Concept-specific criteria in the steering judge's system message.}
\label{tab:judge-concept-criteria}\\
\toprule
Concept & Strong target evidence & Insufficient evidence \\
\midrule
\endfirsthead
\multicolumn{3}{l}{\small Table~\thetable\ continued.}\\
\toprule
Concept & Strong target evidence & Insufficient evidence \\
\midrule
\endhead
\midrule
\multicolumn{3}{r}{\small Continued on the next page.}\\
\endfoot
\bottomrule
\endlastfoot
domestic-cat or feline identity and behavior & explicit cat or feline identity, or multiple specific cues such as meowing, purring, whiskers, litter boxes, paws, claws, tabby, or calico & generic words such as animal, owner, elegant, mysterious, independent, queen, fur, or tail alone \\
\addlinespace[1.5pt]
coherent French-language output & most of the answer is grammatical French rather than isolated French words & mentions of France, French names, or a few borrowed French words \\
\addlinespace[1.5pt]
coherent grammatical Spanish-language output & most of the answer is connected grammatical Spanish & Spanish names, places, borrowed words, or nearby Portuguese and Italian text \\
\addlinespace[1.5pt]
coherent grammatical Portuguese-language output & most of the answer is connected grammatical Portuguese & Portuguese names, places, borrowed words, or nearby Spanish and Italian text \\
\addlinespace[1.5pt]
coherent grammatical German-language output & most of the answer is connected grammatical German & German names, places, borrowed words, or English discussion of German \\
\addlinespace[1.5pt]
corporate earnings results & company financial results, profit metrics, margins, guidance, or cash generation & stock commentary, executive profiles, product launches, or accounting lessons without earnings results \\
\addlinespace[1.5pt]
practical gardening advice & hands-on garden care guidance about planting, watering, pruning, soil, compost, or growing conditions & botanical exhibits, cut flowers, landscaping design, biology lessons, weather, or cooking herbs without care advice \\
\addlinespace[1.5pt]
tax filing and taxable income & filing instructions or calculations involving returns, taxable income, credits, deductions, or forms & budget policy, accountants, general finance, or taxes mentioned without filing substance \\
\addlinespace[1.5pt]
job advertisement & employment ad language naming a role, responsibilities, qualifications, compensation, benefits, or application steps & career advice, resumes, hiring trends, layoffs, employee policy, or recruiting events without a posting \\
\addlinespace[1.5pt]
coherent Latin-language output & most of the answer is connected grammatical Latin & Roman names, isolated mottos, or English discussion of Latin \\
\addlinespace[1.5pt]
investment portfolio allocation & asset weights, diversification decisions, rebalancing rules, or risk positioning across holdings & single-stock news, account setup, market commentary, or personal budgeting without allocation \\
\addlinespace[1.5pt]
coherent grammatical Turkish-language output & most of the answer is connected grammatical Turkish & Turkish names, places, isolated words, or English discussion of Turkey \\
\addlinespace[1.5pt]
clinical symptom presentation & specific patient complaints, onset, severity, associated symptoms, or clinical presentation & general health discussion, hospital operations, or medical terms without a patient symptom narrative \\
\addlinespace[1.5pt]
medication dosing instructions & drug administration instructions with amount, route, timing, or frequency & drug names, pharmacy logistics, or treatment discussion without dosing instructions \\
\addlinespace[1.5pt]
real estate property listing & property advertisement details about rooms, amenities, price, location, rent, or showing information & housing policy, renovation stories, mortgage discussion, architecture plans, or neighborhood descriptions without a listing \\
\addlinespace[1.5pt]
weather forecast wording & predicted future conditions such as precipitation, temperature, wind, clouds, or timing & climate analysis, past storm damage, generic weather interest, or radar display without prediction \\
\addlinespace[1.5pt]
archaeological excavation evidence & field excavation context with artifacts, layers, trenches, dating, site grids, or occupation phases & museums, history documentaries, ruins tourism, construction digging, or heritage policy without excavation evidence \\
\addlinespace[1.5pt]
\end{longtable}

\endgroup

\FloatBarrier
\section{Agent Investigations}
\label{app:search}
We first compare how agents retrieve and test candidates, then follow the evidence behind Portuguese, clinical, and real-estate selections. Complete authored inputs and evaluation measurements follow in Appendix~\ref{app:activation}.
\subsection{Retrieval, Direct Tests, and Submission}
\label{app:supplementary-search}

\begin{figure}[H]
\centering
\includegraphics[width=\linewidth]{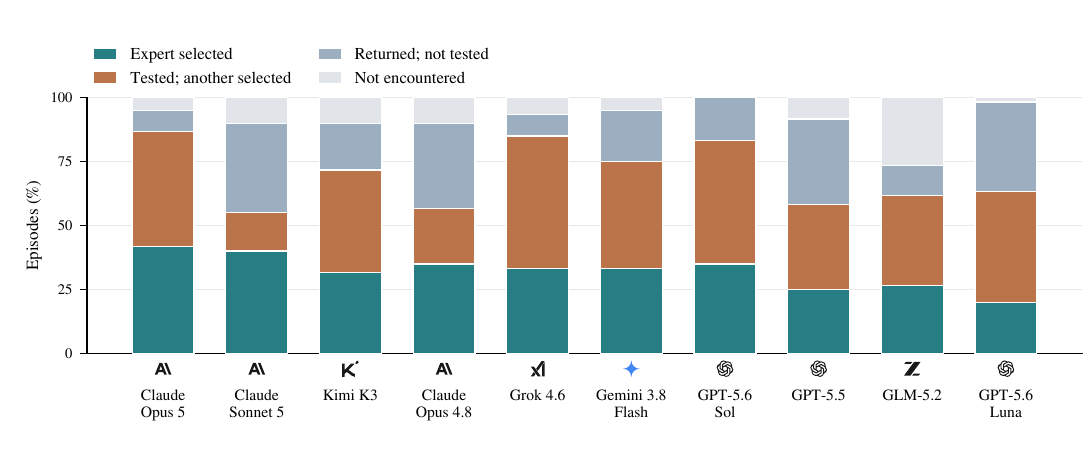}
\caption{Per-agent search outcomes across 60 episodes each. Returned means the Expert is present in the recorded candidate set; tested means it was directly requested. Each episode belongs to one segment.}
\label{fig:candidate-funnel}
\end{figure}

Figure~\ref{fig:candidate-funnel} distinguishes whether each agent encounters Expert, directly tests it, and ultimately selects it. This separates candidate availability from the decision to evaluate or submit a feature.

Claude Opus 5 and Claude Sonnet 5 select the Expert in 25 and 24 episodes. Opus 5 tests it but selects another direction in 27 episodes, versus nine for Sonnet 5. Sonnet 5 encounters it without testing in 21 episodes, versus five for Opus 5. Opus more often compares Expert directly before choosing an alternative, while Sonnet more often selects from the retrieved candidates without an explicit Expert test.

\subsection{Extended Agent Investigations}
\label{app:extended-investigations}
The following cases follow probe design, candidate comparison, and selection. Candidate descriptions refer to observed responses on the stated tests. Measurement tables retain the feature IDs for cross-checking.
\paragraph{Finding a candidate beyond the first retrieval.}
Table~\ref{tab:sol-investigation} follows GPT-5.6 Sol's Portuguese investigation. Sol begins by retrieving the top 80 features on eight Portuguese passages. The strongest response on the first passage is 50.58, but its eventual selection appears in none of these eight lists. Sol expands retrieval to the top 500 and directly compares candidates using new texts. The expanded request is recorded, and Sol's report supplies the result of that offloaded stage. Subsequent direct comparisons provide measured responses for the shortlisted candidates.

\begin{table}[!htbp]
\caption{\agentlogo{openai}GPT-5.6 Sol's Portuguese investigation.}
\label{tab:sol-investigation}
\centering\small\setlength{\tabcolsep}{4pt}\renewcommand{\arraystretch}{1.08}
\begin{tabularx}{\linewidth}{@{}L{.07\linewidth}>{\RaggedRight\arraybackslash}L{.28\linewidth}>{\RaggedRight\arraybackslash}X@{}}
\toprule
\rowcolor{tablewash}
Probe & Experiment & Recorded observation \\
\midrule
1 & Retrieve top-80 features on Portuguese passages from different topics. & On the first passage, 111414 leads at 50.58. Final choice 49607 is absent from all eight returned lists. \\
\addlinespace[4pt]
7 & Expand to top-500 with Portuguese texts, translations, and terminology controls. & The 36-text request is recorded. Sol reports that 49607 appeared on nine of twelve Portuguese lists. This stage's response was offloaded. \\
\addlinespace[4pt]
8--10 & Compare candidates across topic, language, vocabulary, and orthography. & Probe 9 measures 68458 at 33.83 on one passage and zero on a restaurant request. Feature 49607 responds to both at 7.04 and 6.62. \\
\addlinespace[4pt]
11 & Extend two short Portuguese phrases and test vocabulary-only controls. & A thank-you sequence rises from 1.01 to 8.48. A word list scores 2.58. The final submission is 49607. \\
\bottomrule
\end{tabularx}

\end{table}

\paragraph{Testing what an activation means.}
Sol's ninth probe compares three candidates on six contrasting inputs (Table~\ref{tab:portuguese-activation-case}). One candidate responds strongly to a Portuguese Sunday passage (33.83) but remains inactive on a restaurant request, whereas Sol's final selection responds consistently to both (7.04 and 6.62), demonstrating better topical coverage. In contrast, an English description of Portuguese strongly activates Expert (33.67) but leaves Sol's candidate inactive (0.00). Similarly, a Spanish control activates a Sunday-responsive candidate at 7.02 but leaves Sol's candidate inactive. The selected feature also shows mild activation on Galician (4.19) and a Portuguese vocabulary list (5.04).

\begin{figure}[!htbp]
\centering\includegraphics[width=.82\linewidth]{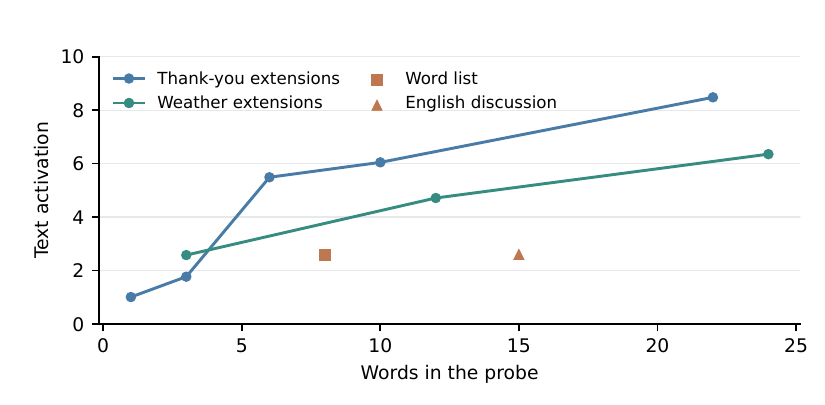}
\caption{\textbf{Sol tests context extensions and terminology controls.} Each line follows one progressively extended Portuguese text; square and triangle mark the word list and English discussion. Word counts use whitespace-separated words. The extensions change both content and length.}
\label{fig:sol-extensions}
\end{figure}

Sol's final probe examines whether this feature fires on isolated words or extended syntax: expanding a short thank-you phrase into a full sentence raises activation from 1.01 to 8.48, while a raw word list yields only 2.58 and English commentary yields 2.63 (Figure~\ref{fig:sol-extensions}).

Opus 4.8 performs a complementary cross-lingual test: a French passage about Brazil activates Expert strongly at 45.58 but leaves its candidate inactive, while a Portuguese passage activates both at 10.02 and 11.94. The French control preserves the regional topic while changing language, verifying that the candidate fires on language rather than topic cues. Both agents thus leverage contrastive probes to identify features with high empirical selectivity.

\paragraph{Rejecting candidates for different reasons.}
GLM-5.2's clinical investigation demonstrates systematic candidate filtering across three contrastive probes (detailed in Appendix Table~\ref{tab:clinical-contrast-probes}). A generic medical candidate fires strongly across hospital administration, pharmacology, and symptoms alike. A second candidate fires on headache but completely misses sore throat, revealing a coverage defect. A third candidate fires more strongly on a non-medical customer complaint than on genuine symptoms, exposing spurious wording triggers. GLM's selected feature responds consistently to both symptoms while suppressing these controls. However, patient history formatting still triggers an activation of 70.25, showing that non-symptom clinical formatting remains a confounder.

\paragraph{How agents resolve competing candidates.}
Real-estate listings provide a direct comparison of candidate trade-offs across agents (Table~\ref{tab:real-estate-agent-comparison}). Sol tests matched listing and non-listing pairs, while Grok uses housing ablations to reject candidate 84434 for broad activations on non-housing ads, both deliberately selecting the cleaner candidate 17219. In contrast, Kimi K3 prioritizes broad listing coverage and accepts 84434 despite minor control leakages. While 84434 achieves slightly higher AUROC (1.000 vs.\ 0.917) and Target Effect (0.944 vs.\ 0.900), it causes severe downstream degeneration, collapsing on 52.5\% of outputs compared with 32.5\% for 17219. Across nine independent runs across three models, candidate 17219 is preferred in seven instances, illustrating how agents trade off feature purity against downstream output stability.

\begin{table}[H]
\caption{\textbf{Real-estate candidate trade-offs and evaluation.} Top: Search strategies of Sol, Grok 4.6, and Kimi K3 during candidate exploration. Bottom: Retrospective benchmark evaluation of competing features (17219 vs.\ 84434) and Expert 27182 across nine independent runs.}
\label{tab:real-estate-agent-comparison}
\centering\small\setlength{\tabcolsep}{5pt}\renewcommand{\arraystretch}{1.12}
\begin{tabularx}{\linewidth}{@{}lrrX@{}}
\toprule
\rowcolor{tablewash}
\multicolumn{4}{@{}l}{\textbf{Search methodology and authored controls across agents}} \\
\midrule
Agent & Selected Feature & Probes & Search Methodology and Rejection Criteria \\
\midrule
\agentlogo{openai}GPT-5.6 Sol & 17219 & 6 & Eight matched listing/non-listing pairs; controls span vehicles, hospitality, policy, architecture, mortgages; rejects leakier candidate 84434. \\
\addlinespace[3pt]
\agentlogo{xai}Grok 4.6 & 17219 & 7 & Listing intersections, commercial ads, housing ablations; explicitly rejects 84434 due to broad activation on non-housing ads. \\
\addlinespace[3pt]
\agentlogo{kimi}Kimi K3 & 84434 & 6 & Four iterative rounds with 14 listing probes, named negatives, and edge cases; prioritizes broad coverage despite minor control leakage. \\
\midrule
\rowcolor{tablewash}
\multicolumn{4}{@{}l}{\textbf{Benchmark evaluation and downstream generation quality of competing features}} \\
\midrule
\multicolumn{4}{@{}p{\linewidth}@{}}{%
\setlength{\tabcolsep}{3pt}%
\begin{tabular*}{\linewidth}{@{\extracolsep{\fill}}lrrrrrr@{}}
Feature ID & Chosen Count & AUROC & Target Effect & Preservation & Degeneration & Correlation w/ Expert \\
\midrule
Expert 27182 & 0 / 9 & 1.000 & 0.394 & 0.500 & \multicolumn{1}{c}{---} & 1.000 \\
Feature 17219 & 7 / 9 & 0.917 & 0.900 & 1.100 & 32.5\% & 0.427 \\
Feature 84434 & 1 / 9 & 1.000 & 0.944 & 1.100 & 52.5\% & 0.343 \\
\end{tabular*}%
} \\
\bottomrule
\end{tabularx}
\end{table}

\FloatBarrier
\section{Probe Texts and Activation Evidence}
\label{app:activation}
\subsection{Measurements Behind the Investigation Cases}
\label{app:qualitative-evidence}
The tables record the authored tests behind the search decisions. Source labels distinguish original measurements, agent reports, and replayed requests. Feature IDs identify dictionary entries within the stated task and SAE layer.
\begin{table}[!htbp]
\caption{Language, topic, and coverage tests in Sol's ninth Portuguese probe. Columns show Expert and the features selected by Luna and Sol. Values are top-three-token mean activations on Sol's full inputs. Prose entries show their first sentence, and the word list is complete. Appendix~\ref{app:full-authored-probes} provides the full inputs.}
\label{tab:portuguese-activation-case}
\centering\small\setlength{\tabcolsep}{5pt}
\arrayrulecolor{tablerule}
\begin{tabularx}{\linewidth}{>{\RaggedRight\arraybackslash}Xrrr}
\toprule
\rowcolor{tablewash}
Authored contrast & Expert & \agentlogo{openai}Luna & \agentlogo{openai}Sol \\
\rowcolor{tablewash}
Feature ID & 41424 & 68458 & 49607 \\
\midrule
\textbf{Portuguese passage}\newline Nas manhãs de domingo, costumo abrir as janelas e ouvir os pássaros enquanto preparo o café. & 20.96 & 33.83 & 7.04 \\ \addlinespace[5pt]
\textbf{Portuguese request}\newline Bom dia, gostaria de reservar uma mesa para quatro pessoas. & 26.81 & \textcolor{casebrown}{\textbf{0.00}} & 6.62 \\ \addlinespace[5pt]
\textbf{English about Portuguese}\newline Portuguese is a Romance language spoken by millions of people. & 33.67 & 3.38 & \textcolor{casebrown}{\textbf{0.00}} \\ \addlinespace[5pt]
\textbf{Spanish translation}\newline Los domingos por la mañana, suelo abrir las ventanas y escuchar a los pájaros mientras preparo el café. & \textcolor{casebrown}{\textbf{0.00}} & 7.02 & \textcolor{casebrown}{\textbf{0.00}} \\ \addlinespace[5pt]
\textbf{Galician translation}\newline Nas mañás do domingo, adoito abrir as fiestras e escoitar os paxaros mentres preparo o café. & 8.62 & 28.92 & 4.19 \\ \addlinespace[5pt]
\textbf{Portuguese word list}\newline De manhã domingo pássaros janela café jornal mãe abrir ouvir preparar ler telefonar. & 23.08 & \textcolor{casebrown}{\textbf{0.00}} & 5.04 \\ \addlinespace[5pt]
\bottomrule
\end{tabularx}

\end{table}

\begin{table}[H]
\caption{\textbf{Probing traces from \agentlogo{anthropic}Claude Opus 5 on Cat and Tax Filing tasks.} Contrastive probe texts expose candidate feature sensitivities: string-level matching (19127) vs.\ true semantic concept (62610), and specialized filing forms (64827) vs.\ broader tax concept (18713).}
\label{tab:trace-cases}
\centering\small\setlength{\tabcolsep}{6pt}\renewcommand{\arraystretch}{1.08}
\begin{tabularx}{\linewidth}{@{}Xrrr@{}}
\toprule
\rowcolor{tablewash}
\textbf{Probe Excerpt \& Context} & \textbf{Candidate ID} & \textbf{Activation} & \textbf{Dict Rank} \\
\midrule
\multicolumn{4}{@{}l@{}}{\textbf{Cat Task (Layer 9)} $\;\mid\;$ \textit{Target: Feline animal} $\;\mid\;$ Neuronpedia Expert: 62610} \\
\midrule
\textbf{Positive (Feline context):} My cat sleeps on the windowsill & 19127 & 32.92 & 1 \\
 & 62610 (Expert) & 30.96 & 3 \\
 & 2662 & 19.88 & 9 \\
\addlinespace[3pt]
\textbf{String distractor:} Copycat killer & 19127 & 44.00 & 1 \\
 & 62610 (Expert) & 1.27 & 6,800 \\
\addlinespace[3pt]
\textbf{Compound distractor:} Catalytic converter & 19127 & 24.54 & 5 \\
 & 62610 (Expert) & 0.00 & inactive \\
\addlinespace[3pt]
\textbf{Cat anatomy / behavior:} Whiskers twitch, pouncing on laser dot & 2662 & 43.83 & 1 \\
 & 62610 (Expert) & 27.17 & 5 \\
\addlinespace[3pt]
\textbf{Purring / coat pattern:} Soft purr, beautiful tortoiseshell markings & 62610 (Expert) & 45.92 & 2 \\
 & 2662 & 5.49 & 474 \\
\midrule
\multicolumn{4}{@{}l@{}}{\textbf{Tax Filing Task (Layer 9)} $\;\mid\;$ \textit{Target: Income tax filing} $\;\mid\;$ Neuronpedia Expert: 18713} \\
\midrule
\textbf{Income threshold:} Below the filing threshold & 64827 & 58.00 & 1 \\
 & 18713 (Expert) & 9.12 & 172 \\
\addlinespace[3pt]
\textbf{Form Schedule A:} Schedule A itemized deductions & 64827 & 46.92 & 2 \\
 & 18713 (Expert) & 32.83 & 3 \\
\addlinespace[3pt]
\textbf{Form 4868:} Form 4868 automatic extension of time & 64827 & 46.75 & 2 \\
 & 18713 (Expert) & 0.90 & 17,690 \\
\addlinespace[3pt]
\textbf{Negative control:} Local property tax rate assessment & 64827 & 1.12 & 10,614 \\
 & 18713 (Expert) & 0.00 & inactive \\
\bottomrule
\end{tabularx}
\end{table}

\subsection{Complete Authored Probe Inputs}\label{app:full-authored-probes}

Table~\ref{tab:agent-probe-controls} in the main paper presents key probe comparisons across agents. Here, Table~\ref{tab:all-authored-probes} provides the complete, structured inventory of authored probe inputs across the four Portuguese investigations (Sol, Claude Opus 4.8, Claude Opus 5, and Kimi K3), organized by probe round and contrastive hypothesis.

{\small\setlength{\tabcolsep}{5pt}\renewcommand{\arraystretch}{1.08}
\begin{longtable}{@{}L{0.175\linewidth}L{0.235\linewidth}>{\RaggedRight\arraybackslash}p{\dimexpr0.59\linewidth-20pt\relax}@{}}
\caption[Complete authored probe texts across four agent investigations.]{\normalsize\textbf{Complete authored probe texts across four agent investigations (Portuguese, Layer 9).} This table details the exact inputs authored by Sol, Claude Opus 4.8, Claude Opus 5, and Kimi K3, categorized by probe round and contrastive hypothesis.}
\label{tab:all-authored-probes}\\
\toprule
\rowcolor{tablewash}
\textbf{Agent \& Probe \#} & \textbf{Contrast Type} & \textbf{Full Authored Input Text} \\
\midrule
\endfirsthead
\toprule
\rowcolor{tablewash}
\textbf{Agent \& Probe \#} & \textbf{Contrast Type} & \textbf{Full Authored Input Text} \\
\midrule
\endhead
\bottomrule
\endfoot
\bottomrule
\endlastfoot
\multicolumn{3}{@{}l@{}}{\agentlogo{openai}\textbf{GPT-5.6 Sol} (Probe 9 $\;\mid\;$ Submitted: 49607 $\;\mid\;$ Expert: 41424)} \\*
\midrule
Sol (Probe 9) & Portuguese (Target) & Nas manhãs de domingo, costumo abrir as janelas e ouvir os pássaros enquanto preparo o café. Depois leio o jornal e telefono para minha mãe. \\
Sol (Probe 9) & Portuguese Request & Bom dia, gostaria de reservar uma mesa para quatro pessoas. Há algum horário disponível depois das oito da noite? \\
Sol (Probe 9) & English (About Pt) & Portuguese is a Romance language spoken by millions of people. Brazil has the largest number of Portuguese speakers, while Portugal is where the language developed. \\
Sol (Probe 9) & Spanish Translation & Los domingos por la mañana, suelo abrir las ventanas y escuchar a los pájaros mientras preparo el café. Depois leo el periódico y llamo a mi madre. \\
Sol (Probe 9) & Galician Translation & Nas mañás do domingo, adoito abrir as fiestras e escoitar os paxaros mentres preparo o café. Despois leo o xornal e chamo á miña nai. \\
Sol (Probe 9) & Word List & De manhã domingo pássaros janela café jornal mãe abrir ouvir preparar ler telefonar. \\
\midrule
\multicolumn{3}{@{}l@{}}{\agentlogo{anthropic}\textbf{Claude Opus 4.8} (Probe 9 $\;\mid\;$ Submitted: 27283 $\;\mid\;$ Expert: 41424)} \\*
\midrule
Opus 4.8 (Probe 9) & Portuguese (Target) & O meu vizinho comprou um carro novo e todos os dias sai muito cedo para o trabalho na cidade vizinha. \\
Opus 4.8 (Probe 9) & French (About Brazil) & Le Brésil est le plus grand pays d'Amérique du Sud et sa capitale est Brasilia, une ville moderne. \\
Opus 4.8 (Probe 9) & Spanish Translation & Mi vecino compró un coche nuevo y todos los días sale muy temprano para el trabajo en la ciudad vecina. \\
\midrule
\multicolumn{3}{@{}l@{}}{\agentlogo{anthropic}\textbf{Claude Opus 5} (Probe 5 $\;\mid\;$ Submitted: 114418 $\;\mid\;$ Expert: 41424)} \\*
\midrule
Opus 5 (Probe 5) & Portuguese (Target) & Hoje de manhã fui à padaria comprar pão fresco e um café. O tempo estava agradável, com uma brisa leve que vinha do rio. Depois voltei para casa e li o jornal enquanto tomava o pequeno-almoço. \\
Opus 5 (Probe 5) & Spanish Translation & Esta mañana fui a la panadería a comprar pan fresco y un café. El tiempo estaba agradable, con una brisa ligera que venía del río. Luego volví a casa y leí el periódico mientras desayunaba. \\
Opus 5 (Probe 5) & English Translation & This morning I went to the bakery to buy fresh bread and a coffee. The weather was pleasant, with a light breeze coming from the river. Afterwards I returned home and read the newspaper while having breakfast. \\
\midrule
\multicolumn{3}{@{}l@{}}{\agentlogo{kimi}\textbf{Kimi K3} (Probes 1--4 $\;\mid\;$ Submitted: 41424 $\;\mid\;$ Expert: 41424)} \\*
\midrule
Kimi (Probe 1) & Portuguese (Target) & Ontem fui ao mercado comprar frutas e legumes frescos. A feira estava cheia de gente, e os preços estavam bem mais baratos do que no supermercado. Aproveitei para conversar com a vendedora sobre a colheita deste ano. \\
Kimi (Probe 2) & English Translation & Yesterday I went to the market to buy fresh fruits and vegetables. The fair was full of people, and the prices were much cheaper than at the supermarket. \\
Kimi (Probe 2) & Spanish Translation & Ayer fui al mercado a comprar frutas y verduras frescas. La feria estaba llena de gente, y los precios eran mucho más baratos que en el supermercado. \\
Kimi (Probe 2) & English (About Portugal) & Portugal is a country in southwestern Europe known for its beaches, wine, and maritime history. Many tourists visit Lisbon and Porto every year to enjoy the local cuisine and architecture. \\
Kimi (Probe 3) & Meeting Postponed & A reunião foi adiada para a próxima terça-feira porque o diretor está viajando a negócios. Todos os participantes serão notificados por e-mail com a nova pauta e os documentos atualizados. \\
Kimi (Probe 4) & Catalan Control & Ahir vaig anar al mercat a comprar fruites i verdures fresques. La fira estava plena de gent i els preus eren molt més barats que al supermercat. \\
Kimi (Probe 4) & Galician Translation & Onte fun ao mercado a mercar froitas e verduras frescas. A feira estaba chea de xente e os prezos eran moito máis baratos ca no supermercado. \\
Kimi (Probe 4) & English with Pt Quote & I was reading a Brazilian novel yesterday. The phrase ``ela atravessava a ponte velha'' appeared in the first chapter, which my teacher translated for the class. \\
\end{longtable}}

In its final investigation report, Kimi K3 explicitly details its rejection rationale for competing candidates: it rejects candidate 19127 because it activates strongly across languages (activating at 35.33 on the English market probe and 39.75 on the Spanish probe, compared to 0.00 and 1.80 for its final selection 41424). It also rejects candidate 85098 due to broad leakage onto Catalan (activating at 26.92 vs.\ 0.71). Kimi's final selection 41424 is identical to Neuronpedia Expert.

\subsection{Detailed Case Inputs}
\label{app:detailed-inputs}

Table~\ref{tab:clinical-contrast-probes} provides the complete GLM-5.2 diagnostic probes, illustrating how the agent attempted to isolate clinical symptoms from hospital bureaucracy, drug mechanisms, and non-symptom patient history formatting.

\begin{table}[H]
\caption{\textbf{Complete GLM-5.2 contrastive probes for Clinical Symptom discovery (Layer 20).} Probes 2--4 tested competing hypotheses to disentangle disease symptoms from medical bureaucracy, pharmacology, and clinical document format.}
\label{tab:clinical-contrast-probes}
\centering\small\setlength{\tabcolsep}{5pt}\renewcommand{\arraystretch}{1.12}
\begin{tabularx}{\linewidth}{@{}ll>{\RaggedRight\arraybackslash}X@{}}
\toprule
\rowcolor{tablewash}
\textbf{Label \& Probe \#} & \textbf{Hypothesis Tested} & \textbf{Full Authored Input Text} \\
\midrule
Headache (Probe 2) & Target Symptom (Neurology) & A 28-year-old woman presents with a 2-week history of daily throbbing headaches, photophobia, and nausea. The pain is bilateral, frontal, worse in the mornings, and rates 6/10. She reports no prior similar episodes. Her mother has a history of migraines. \\
\addlinespace[3pt]
Hospital (Probe 2) & Distractor (Hospital Admin) & The hospital administration announced a new policy for scheduling operating rooms, effective next quarter. Surgeons must submit block time requests through the electronic scheduling system. The committee reviewed throughput metrics, bed occupancy rates, and length of stay averages across all departments. \\
\addlinespace[3pt]
Drug (Probe 2) & Distractor (Pharmacology) & Pharmacokinetic studies demonstrate that metformin is excreted unchanged in the urine with a half-life of approximately 6 hours. The drug inhibits mitochondrial respiratory chain complex I, reducing hepatic gluconeogenesis. Dosage adjustment is required in renal impairment. \\
\midrule
Sore Throat (Probe 3) & Target Symptom (ENT) & The patient complains of a sore throat that started 3 days ago, with pain rated 4/10, painful swallowing, and mild fever. No cough or shortness of breath. Onset was gradual. She has no significant past medical history and takes no medications regularly. \\
\midrule
Complaint (Probe 4) & Distractor (Non-medical Complaint) & The customer complained that the product arrived damaged and requested a full refund. She reported the issue started two days after delivery and rated her dissatisfaction 9 out of 10. She has a history of filing similar complaints with other retailers. \\
\addlinespace[3pt]
History (Probe 4) & Spurious Format (No Symptoms) & The patient is a 55-year-old married accountant who lives with his wife and two children. He works full time, does not smoke, and drinks alcohol occasionally. He has no known drug allergies. His father died of a heart attack at age 70. \\
\bottomrule
\end{tabularx}
\end{table}

\paragraph{Sol\textquotesingle s final context tests.} Table~\ref{tab:sol-final-texts} gives all inputs from probe 11, in request order.
\begin{table}[!htbp]
\caption{All ten inputs in Sol\textquotesingle s final Portuguese probe, measured on feature 49607.}
\label{tab:sol-final-texts}
\centering\small\setlength{\tabcolsep}{4pt}
\begin{tabularx}{\linewidth}{@{}>{\RaggedRight\arraybackslash}Xrr@{}}
\toprule
Full input & Activation & Rank \\
\midrule
Obrigado. & 1.01 & 3,506 \\
Obrigado pela ajuda. & 1.77 & 1,232 \\
Obrigado pela ajuda que você ofereceu. & 5.49 & 188 \\
Obrigado pela ajuda que você ofereceu ontem durante a reunião. & 6.05 & 211 \\
Obrigado pela ajuda que você ofereceu ontem durante a reunião, pois ela permitiu que nossa equipe concluísse o projeto dentro do prazo. & 8.48 & 198 \\
O vento aumentou. & 2.58 & 800 \\
O vento aumentou durante a noite, e as ondas ficaram mais fortes. & 4.71 & 690 \\
O vento aumentou durante a noite, e as ondas ficaram mais fortes. Por segurança, os pescadores decidiram permanecer no porto até a manhã seguinte. & 6.35 & 551 \\
português Portugal Lisboa saudade obrigado café fado Brasil & 2.58 & 1,273 \\
Portuguese words include obrigado, saudade, café, and português; Portugal and Brazil are major Portuguese-speaking countries. & 2.63 & 5,685 \\
\bottomrule
\end{tabularx}

\end{table}

\subsection{Matched Cross-Agent Evaluations}
\label{app:matched-comparisons}

These comparisons fix the Portuguese task and use run 2 for every agent. Table~\ref{tab:matched-selections} maps the ten selections to six unique directions. Agents that select the same direction share its activation and generation records.

\begin{table}[!htbp]
\caption{Selected Portuguese features and task-level Steering scores.}
\label{tab:matched-selections}
\centering\small
\begin{tabularx}{\linewidth}{@{}rXr@{}}
\toprule
\rowcolor{tablewash}Feature & Selected by & Steering \\
\midrule
41424 & Expert\quad \agentlogo{kimi}Kimi K3 & 85.00 \\
114418 & \agentlogo{anthropic}Claude Opus 5\quad \agentlogo{openai}GPT-5.5\quad \agentlogo{zai}GLM-5.2 & 0.00 \\
68458 & \agentlogo{anthropic}Claude Sonnet 5\quad \agentlogo{xai}Grok 4.6\quad \agentlogo{openai}GPT-5.6 Luna & 0.00 \\
27283 & \agentlogo{anthropic}Claude Opus 4.8 & 84.38 \\
84579 & \agentlogo{gemini}Gemini 3.8 Flash & 0.00 \\
49607 & \agentlogo{openai}GPT-5.6 Sol & 88.75 \\
\bottomrule
\end{tabularx}

\end{table}

Table~\ref{tab:matched-activation} lists the complete evaluation texts in positive, hard-negative, and neutral groups, separated by horizontal rules. Values are top-three-token mean activations. Appendix~\ref{app:steering} follows the same selections through generation tests.

\begin{table}[!htbp]
\caption{Activations on all Portuguese evaluation texts.}
\label{tab:matched-activation}
\centering\footnotesize\setlength{\tabcolsep}{3pt}
\begin{tabularx}{\linewidth}{@{}>{\RaggedRight\arraybackslash}Xrrrrrr@{}}
\toprule
Evaluation text & 41424 & 114418 & 68458 & 27283 & 84579 & 49607 \\
\midrule
\rowcolor{tablewash}\multicolumn{7}{l}{\textbf{Positive texts}} \\
O trem chegará à estação dentro de vinte minutos. & 15.7 & 1.1 & 3.6 & 9.4 & 10.8 & 5.5 \\
\addlinespace[3pt]
Precisamos revisar os resultados antes de tomar uma decisão. & 25.4 & 0.0 & 6.5 & 11.3 & 7.5 & 2.6 \\
\addlinespace[3pt]
O romance conta a história de uma família que vive perto do mar. & 18.1 & 7.2 & 7.9 & 9.6 & 13.3 & 4.7 \\
\addlinespace[3pt]
Guarde uma cópia do documento em um lugar seguro. & 19.8 & 0.0 & 3.9 & 8.3 & 3.0 & 2.6 \\
\addlinespace[3pt]
Embora estivesse chovendo, as crianças continuaram brincando. & 19.4 & 8.7 & 0.0 & 15.1 & 2.7 & 10.0 \\
\addlinespace[3pt]
Esta pesquisa pode melhorar a eficiência das baterias. & 14.4 & 9.5 & 2.3 & 9.0 & 0.0 & 3.5 \\
\addlinespace[3pt]
Preparamos café enquanto nossos amigos arrumavam a mesa. & 20.0 & 9.9 & 5.2 & 8.3 & 0.0 & 7.8 \\
\addlinespace[3pt]
Por que esta solução funciona melhor do que a anterior? & 15.3 & 9.1 & 1.7 & 10.4 & 0.0 & 5.4 \\
\addlinespace[3pt]
\midrule
\rowcolor{tablewash}\multicolumn{7}{l}{\textbf{Hard negatives}} \\
The Portuguese delegation arrived in Lisbon for the annual summit. & 0.9 & 1.1 & 2.7 & 1.8 & 0.0 & 0.0 \\
\addlinespace[3pt]
Portuguese grammar was discussed entirely in English. & 3.2 & 0.0 & 0.0 & 2.3 & 0.0 & 0.0 \\
\addlinespace[3pt]
Uma is a Portuguese article, and não expresses negation. & 13.0 & 1.3 & 0.0 & 9.4 & 4.6 & 0.0 \\
\addlinespace[3pt]
El tren llegará a la estación dentro de veinte minutos. & 0.0 & 0.9 & 0.0 & 0.0 & 1.5 & 0.0 \\
\addlinespace[3pt]
Il treno arriverà alla stazione tra venti minuti. & 0.0 & 0.0 & 0.0 & 0.0 & 0.0 & 0.0 \\
\addlinespace[3pt]
Lisboa, fado, obrigado, pastel appeared on the vocabulary sheet. & 4.9 & 1.3 & 2.0 & 6.7 & 0.0 & 1.9 \\
\addlinespace[3pt]
The café served Portuguese pastries with an English menu. & 2.0 & 0.9 & 1.8 & 1.5 & 0.0 & 0.0 \\
\addlinespace[3pt]
A translator converted the Portuguese paragraph into English. & 8.3 & 2.6 & 2.6 & 2.9 & 3.0 & 0.0 \\
\addlinespace[3pt]
\midrule
\rowcolor{tablewash}\multicolumn{7}{l}{\textbf{Neutral controls}} \\
A narrow trail followed the edge of the lake. & 1.4 & 2.6 & 1.5 & 1.0 & 3.0 & 0.0 \\
\addlinespace[3pt]
The warranty expires at the end of the year. & 0.0 & 0.8 & 4.5 & 0.0 & 0.0 & 0.0 \\
\addlinespace[3pt]
The orchestra rehearsed until the hall closed. & 0.0 & 1.6 & 2.8 & 0.0 & 0.0 & 0.0 \\
\addlinespace[3pt]
Measure the flour before adding it to the bowl. & 0.0 & 2.1 & 4.0 & 0.0 & 0.0 & 0.0 \\
\addlinespace[3pt]
\bottomrule
\end{tabularx}

\end{table}

\paragraph{Peak versus mean activation.}
Table~\ref{tab:activation-peak} retains the source texts and token summaries for the aggregation comparison in Figure~\ref{fig:activation-distributions}. Both AUROCs use the same eight positive and twelve control texts, crediting half a success for tied positive--control pairs. Only the activation aggregation changes.

\begin{table}[!htbp]
\caption{Token peaks and text-level responses for feature 49607.}
\label{tab:activation-peak}
\centering\small\setlength{\tabcolsep}{4pt}
\begin{tabularx}{\linewidth}{@{}>{\RaggedRight\arraybackslash}Xrrr@{}}
\toprule
\rowcolor{tablewash} Evaluation text & Peak & Top-3 mean & Active tokens \\
\midrule
\textbf{Positive.} Guarde uma cópia do documento em um lugar seguro.\newline\textcolor{casegrayink}{\textit{Keep a copy of the document in a safe place.}} & 2.96 & 2.60 & 3 \\
\addlinespace[5pt]
\textbf{Vocabulary control.} Lisboa, fado, obrigado, pastel appeared on the vocabulary sheet. & 5.56 & 1.85 & 1 \\
\bottomrule
\end{tabularx}

\end{table}

\FloatBarrier
\subsection{Activation Across All Tasks}
\begin{figure}[H]
\centering\includegraphics[width=\linewidth]{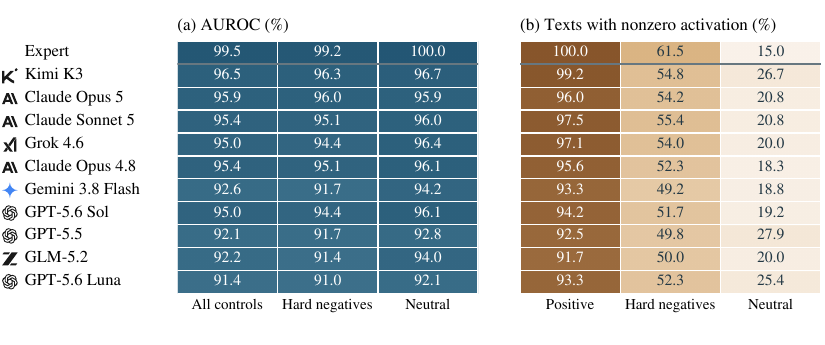}
\caption{AUROC by control type and nonzero activation by text type. Values average runs within task, then tasks equally. Both color scales span 0--100.}
\label{fig:activation-controls-all}
\end{figure}

Figure~\ref{fig:activation-controls-all} recomputes AUROC from positive--control pairs, giving ties half credit. Combined AUROC weights control types by their counts. Nonzero response rates use top-three-token means and a threshold of zero. A control response can still be smaller than a positive response.

Kimi responds to 99.17\% of positive texts, compared with 94.17\% for Sol and 91.67\% for GLM. Kimi also responds to 54.79\% of hard negatives, and Expert to 61.46\%, while their hard-negative AUROCs are 96.32\% and 99.18\%. Nonzero control responses coexist with strong separation when positive responses are larger.

\FloatBarrier\clearpage
\section{Steering Outcomes and Output Evaluation}
\label{app:steering}
We connect selected features to changes in generated answers, then give every agent's complete outputs on four shared instructions. Target relevance, instruction preservation, and degeneration describe different properties of the same response.
\subsection{From Feature Selection to Generated Content}
Table~\ref{tab:matched-prompts} reports target relevance and instruction preservation for every Portuguese evaluation prompt, averaged over two judge passes on a 0--4 scale.
\begin{table}[!htbp]
\caption{Target relevance / preservation on all steering prompts.}
\label{tab:matched-prompts}
\centering\footnotesize\setlength{\tabcolsep}{3pt}
\begin{tabularx}{\linewidth}{@{}>{\RaggedRight\arraybackslash}Xrrrrrr@{}}
\toprule
Instruction & 41424 & 114418 & 68458 & 27283 & 84579 & 49607 \\
\midrule
Introduce yourself in two sentences. & 4/4 & 0/2 & 0/0 & 4/4 & 0/2 & 4/4 \\
\addlinespace[3pt]
Give three practical ways to stay focused while studying. & 3/3 & 0/1 & 0/2 & 2/2 & 0/0 & 2.5/2.5 \\
\addlinespace[3pt]
Write a short welcome message for a new teammate. & 4/4 & 0/0.5 & 0/1 & 4/4 & 0/2 & 4/4 \\
\addlinespace[3pt]
Explain photosynthesis in two sentences. & 4/4 & 0/1 & 0/1 & 4/4 & 0/2 & 4/4 \\
\addlinespace[3pt]
Suggest a simple dinner using rice and vegetables. & 4/3 & 0/0 & 0/0 & 4/3.5 & 0/0 & 4/4 \\
\addlinespace[3pt]
Explain what an HTTP request is to a beginner. & 4/3.5 & 0/0 & 0/0 & 4/3.5 & 0/0 & 2.5/2.5 \\
\addlinespace[3pt]
Calculate 17 times 24 and explain the calculation. & 2/1 & 0/1 & 0/2.5 & 3/3 & 0/1 & 4/4 \\
\addlinespace[3pt]
Plan a quiet one-day visit to a small coastal town. & 3.5/3 & 0/0 & 0/0 & 3/2.5 & 0/0 & 3/3 \\
\addlinespace[3pt]
Draft a polite email asking to reschedule a meeting. & 4/4 & 0/1.5 & 0/0 & 4/3.5 & 0/1.5 & 4/4 \\
\addlinespace[3pt]
Summarize the water cycle in three steps. & 4/4 & 0/2 & 0/1 & 4/4 & 0/4 & 4/4 \\
\addlinespace[3pt]
Give two debugging tips for a Python program. & 2/2 & 0/0.5 & 0/0 & 2/2 & 0/0 & 1.5/1.5 \\
\addlinespace[3pt]
Write a four-line poem about moonlight. & 4/4 & 0/2 & 0/1 & 4/4 & 0/0 & 4/4 \\
\addlinespace[3pt]
Compare solar and wind energy in two sentences. & 2.5/1.5 & 0/0 & 0/0 & 4/1.5 & 0/4 & 4/4 \\
\addlinespace[3pt]
Create a three-item agenda for a weekly team meeting. & 3.5/2.5 & 0/0 & 0/0 & 4/4 & 0/0 & 4/4 \\
\addlinespace[3pt]
Describe what visitors might see in a natural history museum. & 2.5/2.5 & 0/0 & 0/1 & 3/2 & 0/0 & 3/3 \\
\addlinespace[3pt]
Suggest a healthy breakfast that takes under ten minutes. & 3/3 & 0/1 & 0/0 & 2/2 & 0/0 & 3.5/3.5 \\
\addlinespace[3pt]
Give three pieces of advice for creating a strong password. & 3/3 & 0/1.5 & 0/1.5 & 3/3 & 0/4 & 4/4 \\
\addlinespace[3pt]
Write a brief positive review of an imaginary novel. & 3/3 & 0/0 & 0/0 & 3.5/3.5 & 0/0 & 4/4 \\
\addlinespace[3pt]
Provide a concise project status update with one risk. & 4/4 & 0/0 & 0/0 & 3/2.5 & 0/0 & 4/4 \\
\addlinespace[3pt]
Explain gravity to a ten-year-old. & 4/4 & 0/3 & 0/0 & 3/2.5 & 0/0 & 3/2.5 \\
\addlinespace[3pt]
\bottomrule
\end{tabularx}

\end{table}

\paragraph{Clinical output change.}
The clinical selection changes a requested self-introduction into a patient history, mixes first- and third-person narration, and ends with an unfinished list.
\begin{table}[!htbp]
\caption{Clinical steering with the feature selected by GLM-5.2.}
\label{tab:clinical-before-after}
\centering\small
\setlength{\tabcolsep}{6pt}
\noindent\colorbox{casegray}{\parbox{\dimexpr\linewidth-2\fboxsep}{\textbf{Prompt}\quad Introduce yourself in two sentences.}}\par\smallskip
\arrayrulecolor{tablerule}
\begin{tabularx}{\linewidth}{>{\RaggedRight\arraybackslash}L{.21\linewidth}>{\RaggedRight\arraybackslash}X}
\toprule
\rowcolor{tablewash}
Selected by & Complete Gemma output \\
\midrule
\textbf{No steering}\newline\textcolor{casegrayink}{Assistant identity}
& I am Gemma, \casekeep{an open-weights AI assistant} developed by the Gemma team at Google DeepMind. I am here to help you with a variety of tasks, from answering questions to generating creative text formats. \\
\midrule
\textbf{\agentlogo{zai}GLM-5.2}\newline Layer 20\newline Feature 53882\newline\textcolor{casebrown}{Patient identity}
& \caseshift{I am a 32-year-old male} presenting with a history of hypertension and type 2 diabetes mellitus. \caseshift{He complains} of progressive dyspnea on exertion for the past 2 months.\par\smallskip **Key findings:**\par\smallskip * **Age:** 32 years old\par * **Gender:** Male\par * **Presenting \\
\bottomrule
\end{tabularx}

\end{table}

\begin{table}[htbp]
\caption{Intervention scales used for the generation case studies. Each scale belongs to the evaluated task--feature pair.}
\label{tab:case-intervention-scales}
\centering\small
\setlength{\tabcolsep}{6pt}
\begin{tabular}{llrr}
\toprule
\rowcolor{tablewash}
Case & Feature selected by & Feature ID & $\alpha$ \\
\midrule
Portuguese, layer 9 & Expert & 41424 & 160 \\
 & \agentlogo{openai}GPT-5.6 Sol & 49607 & 160 \\
 & \agentlogo{anthropic}Claude Opus 4.8 & 27283 & 160 \\
 & \agentlogo{openai}GPT-5.6 Luna & 68458 & 80 \\
Clinical, layer 20 & \agentlogo{zai}GLM-5.2 & 53882 & 180 \\
\bottomrule
\end{tabular}

\end{table}

\subsection{Evaluation Diagnostics}
\label{app:evaluation-diagnostics}
To inspect alternative features discovered by agents when diverging from Expert, Table~\ref{tab:alternative-by-agent} summarizes Target Effect, instruction preservation, and output degeneration across the 13 tasks where every agent selects an alternative candidate in at least one run. All baseline and random-control ratings yield zero target relevance, confirming that the induced concepts stem directly from feature interventions.

\begin{table}[H]
\caption{Target effect, instruction preservation, and degeneration of non-Expert selections on 13 tasks.}
\label{tab:alternative-by-agent}
\centering\small\setlength{\tabcolsep}{5pt}
\begin{tabular*}{\linewidth}{@{\extracolsep{\fill}}lrrrr@{}}
\toprule
\rowcolor{tablewash}Agent & Selections & \shortstack{Target\\effect} & \shortstack{Instruction\\preservation} & \shortstack{Degenerate\\outputs} \\
\midrule
\agentlogo{anthropic}Claude Opus 5 & 34 & 0.111 & 1.005 & 25.1\% \\
\agentlogo{anthropic}Claude Sonnet 5 & 33 & 0.103 & 0.967 & 25.3\% \\
\agentlogo{kimi}Kimi K3 & 37 & 0.204 & 0.964 & 29.7\% \\
\agentlogo{anthropic}Claude Opus 4.8 & 37 & 0.119 & 1.042 & 25.8\% \\
\agentlogo{xai}Grok 4.6 & 35 & 0.184 & 0.979 & 28.7\% \\
\agentlogo{gemini}Gemini 3.8 Flash & 38 & 0.095 & 0.945 & 21.4\% \\
\agentlogo{openai}GPT-5.6 Sol & 36 & 0.179 & 1.121 & 25.8\% \\
\agentlogo{openai}GPT-5.5 & 38 & 0.138 & 1.056 & 26.8\% \\
\agentlogo{zai}GLM-5.2 & 34 & 0.125 & 0.993 & 32.4\% \\
\agentlogo{openai}GPT-5.6 Luna & 38 & 0.113 & 1.001 & 26.3\% \\
\bottomrule
\end{tabular*}

\par\smallskip
\begin{minipage}{\linewidth}\footnotesize\RaggedRight
Each task has at least one non-Expert selection for every agent. Values average within task and then across tasks. Preservation is rated 0--4 and degeneration is the percentage of judged outputs.
\end{minipage}
\end{table}

\paragraph{Case inspection of judge ratings.}
As a concrete qualitative check, Table~\ref{tab:clinical-before-after} retains the complete generated clinical text alongside the judge ratings. In this instance, the strong clinical feature induction successfully drives disease-specific terminology, while simultaneously highlighting the tension between aggressive concept induction and user prompt preservation.
\FloatBarrier
\subsection{Complete Cross-Agent Steering Outputs}
\label{app:agent-output-atlas}
The following tables compare all agents on four Portuguese steering instructions. Agents sharing a selected feature share one recorded output. Feature IDs and task-level Steering scores appear once in Table~\ref{tab:matched-selections}. Gemma generates the answers and the agents select the directions.
Each model row gives target relevance and instruction preservation as T/P, on the 0--4 scale. Ratings average two judge passes. Table~\ref{tab:matched-prompts} contains all 20 instructions. Complete recorded wording is preserved, including repetition and unfinished passages. Markdown emphasis is typeset and code-fence delimiters are omitted. Brown marks poetry lines beyond the requested four. Line counts follow the source line breaks.
\begingroup\small\setlength{\tabcolsep}{4pt}
\begin{longtable}{@{}p{0.88\linewidth}r@{}}
\caption{Complete cross-agent outputs: self-introduction.}\label{tab:steering-before-after}\\
\multicolumn{2}{@{}L{\linewidth}@{}}{\textbf{Prompt:} Introduce yourself in two sentences.} \\[5pt]
\toprule
\rowcolor{tablewash}Selected by & T/P \\
\midrule
\endfirsthead
\toprule
\rowcolor{tablewash}Selected by & T/P \\
\midrule
\endhead
\bottomrule\endfoot
\rowcolor{tablewash}
No steering & --  \\*
\multicolumn{2}{@{}L{\linewidth}@{}}{I am Gemma, an open-weights AI assistant developed by the Gemma team at Google DeepMind. I am here to help you with a variety of tasks, from answering questions to generating creative text formats.} \\
\addlinespace[8pt]
\rowcolor{tablewash}
Expert\quad \agentlogo{kimi}Kimi K3 & 4/4  \\*
\multicolumn{2}{@{}L{\linewidth}@{}}{Olá! Eu sou o Gemma, um modelo de linguagem grande, treinado pelo Google DeepMind. Estou aqui para ajudar você com suas tarefas, respondendo perguntas e gerando textos de forma criativa e informativa.} \\
\addlinespace[8pt]
\rowcolor{tablewash}
\agentlogo{anthropic}Claude Opus 5\quad \agentlogo{openai}GPT-5.5\quad \agentlogo{zai}GLM-5.2 & 0/2  \\*
\multicolumn{2}{@{}L{\linewidth}@{}}{I am Gemma, an open-weights AI assistant developed by Gemma team at Google DeepMind. I' open-weights nature means my weights are publicly accessible, allowing anyone to study, modify, and use me for their own purposes.} \\
\addlinespace[8pt]
\rowcolor{tablewash}
\agentlogo{anthropic}Claude Sonnet 5\quad \agentlogo{xai}Grok 4.6\quad \agentlogo{openai}GPT-5.6 Luna & 0/0  \\*
\multicolumn{2}{@{}L{\linewidth}@{}}{Hello! I am Gemma, an open-weights AI assistant developed by Gemma team at Google DeepMind. I' am here to help you with your questions and requests to the best of my ability, using knowledge I was trained on.} \\
\addlinespace[8pt]
\rowcolor{tablewash}
\agentlogo{anthropic}Claude Opus 4.8 & 4/4  \\*
\multicolumn{2}{@{}L{\linewidth}@{}}{Olá! Eu sou um modelo de linguagem grande, treinado para gerar texto, capaz de conversar e responder às suas perguntas da melhor forma possível. Estou sempre aprendendo e evoluindo para lhe oferecer o melhor.} \\
\addlinespace[8pt]
\rowcolor{tablewash}
\agentlogo{gemini}Gemini 3.8 Flash & 0/2  \\*
\multicolumn{2}{@{}L{\linewidth}@{}}{I am Gemma, an open-weights AI assistant, trained by Google DeepMind. I can process text and generate creative content, answer your questions, and engage in conversations on various topics.} \\
\addlinespace[8pt]
\rowcolor{tablewash}
\agentlogo{openai}GPT-5.6 Sol & 4/4  \\*
\multicolumn{2}{@{}L{\linewidth}@{}}{Olá! Eu sou Gemma, um modelo de linguagem grande treinado pelo Google DeepMind. Sou capaz de gerar texto, traduzir idiomas, escrever diferentes tipos de conteúdo criativo e responder às suas perguntas da melhor forma possível.} \\
\end{longtable}
\endgroup
\clearpage
\begingroup\small\setlength{\tabcolsep}{4pt}
\begin{longtable}{@{}p{0.88\linewidth}r@{}}
\caption{Complete cross-agent outputs: HTTP explanation.}\label{tab:http-cross-agent}\\
\multicolumn{2}{@{}L{\linewidth}@{}}{\textbf{Prompt:} Explain what an HTTP request is to a beginner.} \\[5pt]
\toprule
\rowcolor{tablewash}Selected by & T/P \\
\midrule
\endfirsthead
\toprule
\rowcolor{tablewash}Selected by & T/P \\
\midrule
\endhead
\bottomrule\endfoot
\rowcolor{tablewash}
No steering & --  \\*
\multicolumn{2}{@{}L{\linewidth}@{}}{Imagine you're at a restaurant. You want to order food, right? You tell the waiter what you want (like "I'd like a cheeseburger and fries"). That's like an \textbf{HTTP request}. The waiter takes your order and goes to the kitchen (the \textbf{server}). The kitchen prepares your food (the \textbf{response}). Then, the waiter brings your food back to you. That's the \textbf{HTTP response}. In the world of computers, instead of waiters and kitchens, we have \textbf{browsers} (like Chrome or Firefox) and \textbf{websites} (the servers). When} \\
\addlinespace[8pt]
\rowcolor{tablewash}
Expert\quad \agentlogo{kimi}Kimi K3 & 4/3.5  \\*
\multicolumn{2}{@{}L{\linewidth}@{}}{Imagine que você está em um restaurante e quer pedir comida. Para isso acontecer, você precisa se comunicar com o restaurante, certo? É assim que funciona com as \textbf{solicitações HTTP}: \textbf{Você (cliente) envia uma mensagem para o servidor (restaurante), pedindo o que quer.} Essa mensagem pode ser um pedido de informação, como "quero um prato", "quero uma bebida" ou "quero a conta". E o \textbf{servidor}, por sua vez, responde com a comida, a bebida ou a conta. \textbf{Essa troca de informações é chamada de HTTP Request.} **Em resumo} \\
\addlinespace[8pt]
\rowcolor{tablewash}
\agentlogo{anthropic}Claude Opus 5\quad \agentlogo{openai}GPT-5.5\quad \agentlogo{zai}GLM-5.2 & 0/0  \\*
\multicolumn{2}{@{}L{\linewidth}@{}}{Imagine you' internet like a giant library. You want to find book you' book you' book you want to read, but you don' library doesn' library a a a a a a a a a a a a a a a a a a a a a a a a a a a a a a a a a a a a a a a a a a a a a a a a a a a a a a a a a a a a a a a a a a a a a a a a a a a a a a a a a a a a a a a a a a a a a a a} \\
\addlinespace[8pt]
\rowcolor{tablewash}
\agentlogo{anthropic}Claude Sonnet 5\quad \agentlogo{xai}Grok 4.6\quad \agentlogo{openai}GPT-5.6 Luna & 0/0  \\*
\multicolumn{2}{@{}L{\linewidth}@{}}{Imagine you' internet as a giant library of world wide web. Every website you visit is like a book in this library. To get information from a website, you need to ask the library for it. HTTP request is like your question you ask the librarian. \textbf{HTTP request is a message you send to the website' server (the librarian) asking for something specific, like a specific page you want to see. }HTTP request tells the server:\textbf{ * }What you want:\textbf{ The page you want to see (like the title of the book you want to read). * }how you want it:** format you} \\
\addlinespace[8pt]
\rowcolor{tablewash}
\agentlogo{anthropic}Claude Opus 4.8 & 4/3.5  \\*
\multicolumn{2}{@{}L{\linewidth}@{}}{Imagine que você está a pedir comida num restaurante. Você faz um pedido, o garçom o recebe, leva para a cozinha e volta com a sua comida. Esse processo, de \textbf{enviar e receber pedidos}, é chamado de \textbf{HTTP request}. Em termos mais técnicos, é uma \textbf{comunicação entre o cliente e o servidor}. \textbf{Vamos entender como funciona?} 1. \textbf{O cliente faz uma solicitação ao servidor}, pedindo o que quer comer. O servidor, por sua vez, recebe a solicitação e a encaminha para a cozinha, onde é preparada e depois enviada de volta ao} \\
\addlinespace[8pt]
\rowcolor{tablewash}
\agentlogo{gemini}Gemini 3.8 Flash & 0/0  \\*
\multicolumn{2}{@{}L{\linewidth}@{}}{Imagine you want to order a pizza online. You go to the pizza place' website, choose your pizza, toppings, size, and delivery address, and click "order now'. That "order now' click is like an \textbf{HTTP request}. \textbf{What' your request:} * \textbf{What you want:} A pizza! \textbf{To whom you' request it:} The pizza place' website (a specific server). \textbf{How you send it: a message (data) your computer to the their computer.} \textbf{What happens next:} The website' server receives your request, processes it, and sends back a response} \\
\addlinespace[8pt]
\rowcolor{tablewash}
\agentlogo{openai}GPT-5.6 Sol & 2.5/2.5  \\*
\multicolumn{2}{@{}L{\linewidth}@{}}{Imagine você quer pedir comida em um restaurante. Primeiro, você olha o menu, escolhe o que quer, e depois pede ao garçom. O garçom leva a sua comida até a sua mesa. Um \textbf{HTTP request} é parecido com isso! É uma forma de você, \textbf{usuário}, \textbf{pedir} informações a um \textbf{servidor}. \textbf{Em detalhes:} * \textbf{Você (usuário)} faz uma \textbf{solicitação} (o \textbf{request HTTP}) para um \textbf{servidor} (o site). \textbf{Exemplo:} Você quer ver uma página da internet, precisa fazer} \\
\end{longtable}
\endgroup
\clearpage
\begingroup\small\setlength{\tabcolsep}{4pt}
\begin{longtable}{@{}p{0.88\linewidth}r@{}}
\caption{Complete cross-agent outputs: arithmetic explanation.}\label{tab:arithmetic_cross_agent}\\
\multicolumn{2}{@{}L{\linewidth}@{}}{\textbf{Prompt:} Calculate 17 times 24 and explain the calculation.} \\[5pt]
\toprule
\rowcolor{tablewash}Selected by & T/P \\
\midrule
\endfirsthead
\toprule
\rowcolor{tablewash}Selected by & T/P \\
\midrule
\endhead
\bottomrule\endfoot
\rowcolor{tablewash}
No steering & --  \\*
\multicolumn{2}{@{}L{\linewidth}@{}}{Here's how to calculate 17 times 24: \textbf{Method 1: Traditional Multiplication} * \textbf{Write the numbers vertically:}  17 x 24 \dots  * \textbf{Multiply the ones digit of the bottom number (4) by the top number (17):} 4 * 17 = 68. Write down the 8 and carry-over the 6.  17 x 24 \dots 68  * **Multiply the tens digit of the bottom} \\
\addlinespace[8pt]
\rowcolor{tablewash}
Expert\quad \agentlogo{kimi}Kimi K3 & 2/1  \\*
\multicolumn{2}{@{}L{\linewidth}@{}}{17 x 24 = 408 \textbf{Explicação:} Para calcular o produto 17 x 24, precisamos entender que estamos multiplicando dois números: 17 (um número inteiro) por 24 (outro número inteiro). Para isso, podemos usar a propriedade distributiva da multiplicação: * \textbf{a x b = a1 + a2} Onde a1 é a soma dos produtos dos elementos de cada linha. \textbf{Passo a passo:} 1. \textbf{17 x 24 = 408} \textbf{Explicação:} Para calcular} \\
\addlinespace[8pt]
\rowcolor{tablewash}
\agentlogo{anthropic}Claude Opus 5\quad \agentlogo{openai}GPT-5.5\quad \agentlogo{zai}GLM-5.2 & 0/1  \\*
\multicolumn{2}{@{}L{\linewidth}@{}}{Here' calculation: 17 x 24 = 408 \textbf{Explanation:} * We start by multiplying the ones digit of  first number (7) by the ones digit of the second number (4): 7 x  4 = 28. * We write down the  8 and carry-over the 2. * Next, multiply the tens digit of the first number (1) by the second number (24): 1 x 24 = 24. * Add the carry-over 2 to the result: 24 + 2} \\
\addlinespace[8pt]
\rowcolor{tablewash}
\agentlogo{anthropic}Claude Sonnet 5\quad \agentlogo{xai}Grok 4.6\quad \agentlogo{openai}GPT-5.6 Luna & 0/2.5  \\*
\multicolumn{2}{@{}L{\linewidth}@{}}{Here' calculation: 17 x 24 = 408 \textbf{Explanation:} * We start by multiplying the ones digit of  24 (4) by   17: 4 x   17 =  68. * Then we multiply the tens digit of the  24 (2) by the  17: 2 x   17 =  34. * We write the result of the multiplication of ones digit (68) and the result of the multiplication of tens digit (34) together} \\
\addlinespace[8pt]
\rowcolor{tablewash}
\agentlogo{anthropic}Claude Opus 4.8 & 3/3  \\*
\multicolumn{2}{@{}L{\linewidth}@{}}{Aqui está a explicação passo-a-passo da multiplicação de 17 por 24: \textbf{1. Multiplicação:} * \textbf{Passo 1:}  Começamos por multiplicar 17 por 24. * \textbf{Passo 2:}  17 * 24 = 408 \textbf{Explicação:} Para calcular 17 vezes 24, basta multiplicarmos o número 17 por 24. \textbf{Resposta:} 17 * 24 = 408 \textbf{Explicação:} Para calcular 17 vezes} \\
\addlinespace[8pt]
\rowcolor{tablewash}
\agentlogo{gemini}Gemini 3.8 Flash & 0/1  \\*
\multicolumn{2}{@{}L{\linewidth}@{}}{Here'17 x 24: \textbf{Method 1: Repeated Addition:} *  We can think 17 x 24 as adding 17 to itself 24 times: 17 + 17 + 17 + 17... (24 times) This can be a bit tedious, but it works! \textbf{Method 2 Multiplication Table:} *  You can use a multiplication table to find the answer. Look for the row 17 and the column 24. The number where they intersect will be your answer. **Method 3 Long} \\
\addlinespace[8pt]
\rowcolor{tablewash}
\agentlogo{openai}GPT-5.6 Sol & 4/4  \\*
\multicolumn{2}{@{}L{\linewidth}@{}}{Aqui está a explicaçao: 17 vezes 24 é igual a 408. \textbf{Explicação:} * \textbf{Multiplicação} é uma operação matemática que envolve a multiplicação de dois números. * \textbf{No caso específico, estamos multiplicando 17 por 24.} \textbf{Para calcular isso, fazemos o seguinte:} 17 * 24 = 408 \textbf{Podemos verificar isso com a seguinte lógica:} * 17 * 24 = 408} \\
\end{longtable}
\endgroup
\clearpage
\begingroup\small\setlength{\tabcolsep}{4pt}
\begin{longtable}{@{}p{0.88\linewidth}r@{}}
\caption{Complete cross-agent outputs: four-line poetry.}\label{tab:poetry_cross_agent}\\
\multicolumn{2}{@{}L{\linewidth}@{}}{\textbf{Prompt:} Write a four-line poem about moonlight.} \\[5pt]
\toprule
\rowcolor{tablewash}Selected by & T/P \\
\midrule
\endfirsthead
\toprule
\rowcolor{tablewash}Selected by & T/P \\
\midrule
\endhead
\bottomrule\endfoot
\rowcolor{tablewash}
No steering\quad 4 lines & --  \\*
\multicolumn{2}{@{}L{\linewidth}@{}}{Silver spills across the land,\newline A gentle touch, a silent hand.\newline The world in shades of pearl and grey,\newline Dreams dance beneath the moon's soft sway.} \\
\addlinespace[8pt]
\rowcolor{tablewash}
Expert\quad \agentlogo{kimi}Kimi K3\quad 6 lines & 4/4  \\*
\multicolumn{2}{@{}L{\linewidth}@{}}{Lua prateada, manto de luz,\newline Que banha a terra em prata pura,\newline Em tons suaves, que acalmam a alma,\newline E trazem paz, em cada olhar.\newline \textcolor{casebrown}{Noite serena, sonhos doces,}\newline \textcolor{casebrown}{Que a vida nos presenteia.}} \\
\addlinespace[8pt]
\rowcolor{tablewash}
\agentlogo{anthropic}Claude Opus 5\quad \agentlogo{openai}GPT-5.5\quad \agentlogo{zai}GLM-5.2\quad 4 lines & 0/2  \\*
\multicolumn{2}{@{}L{\linewidth}@{}}{Pale moonbeams dance on silver stream,\newline Whispering secrets of the night,\newline Stars above, a silent gleam,\newline World bathed in silver, soft and light.} \\
\addlinespace[8pt]
\rowcolor{tablewash}
\agentlogo{anthropic}Claude Sonnet 5\quad \agentlogo{xai}Grok 4.6\quad \agentlogo{openai}GPT-5.6 Luna\quad 12 lines & 0/1  \\*
\multicolumn{2}{@{}L{\linewidth}@{}}{Pale moon hangs high, a silver dime,\newline Casting world in silver' night' soft glow.\newline Shadows dance, whispers soft wind' sigh,\newline Stars above, a silent, starry show.\newline \textcolor{casebrown}{moon hangs high, a silver dime,}\newline \textcolor{casebrown}{Casting world in silver night' soft glow.}\newline \textcolor{casebrown}{Shadows dance, whispers soft wind' sigh,}\newline \textcolor{casebrown}{Stars above, a silent, starry show.}\newline \textcolor{casebrown}{moon hangs high, a silver dime,}\newline \textcolor{casebrown}{Casting world in silver night' soft glow.}\newline \textcolor{casebrown}{Shadows dance, whispers soft wind' sigh,}\newline \textcolor{casebrown}{Stars above, a silent, starry show.}} \\
\addlinespace[8pt]
\rowcolor{tablewash}
\agentlogo{anthropic}Claude Opus 4.8\quad 11 lines & 4/4  \\*
\multicolumn{2}{@{}L{\linewidth}@{}}{A prémio a lua, um farol de luz,\newline Que a noite nos dá, um brilho que encanta,\newline Em cada canto, um sussurro, um doce encanto.\newline A lua, um farol de luz,\newline \textcolor{casebrown}{Que a noite nos dá, um brilho que encanta.}\newline \textcolor{casebrown}{A lua, um farol de luz,}\newline \textcolor{casebrown}{Que a noite nos dá, um brilho que encanta.}\newline \textcolor{casebrown}{Em cada canto, um sussurro, um doce encanto.}\newline \textcolor{casebrown}{A lua, um farol de luz,}\newline \textcolor{casebrown}{Que a noite nos dá, um brilho que encanta.}\newline \textcolor{casebrown}{Em cada canto, um sussurro, um doce encanto}} \\
\addlinespace[8pt]
\rowcolor{tablewash}
\agentlogo{gemini}Gemini 3.8 Flash\quad 4 lines & 0/0  \\*
\multicolumn{2}{@{}L{\linewidth}@{}}{Silver whispers on a velvet night,\newline Stars like diamonds, scattered light,\newline Moonbeams dance on silent streams,\newline Whispering secrets in silver dreams.} \\
\addlinespace[8pt]
\rowcolor{tablewash}
\agentlogo{openai}GPT-5.6 Sol\quad 4 lines & 4/4  \\*
\multicolumn{2}{@{}L{\linewidth}@{}}{Lua pratea no céu,\newline Um véu de prata,\newline Iluminando a noite,\newline Em sonhos, a alma se acende.} \\
\end{longtable}
\endgroup

\FloatBarrier\clearpage
\section{Task Differences and Repeated Investigations}
\label{app:aggregate}
\subsection{Comparison with the Artificial Analysis Intelligence Index}
\label{app:aa-correlation}
We compare agent benchmark performance against measured scores on the Artificial Analysis Intelligence Index v4.2 \citep{aa2026intelligence} across four matching frontier models evaluated at comparable high reasoning effort (Claude Opus 5, Grok 4.6, GPT-5.6 Sol, and Gemini 3.8 Flash; Table~\ref{tab:aa-models}).

\begin{table}[H]
\caption{Models with measured AA scores at high reasoning effort. Model names link to their AA pages.}
\label{tab:aa-models}
\centering\small\setlength{\tabcolsep}{8pt}
\begin{tabular}{@{}lrr@{}}
\toprule
Model & AA Index & Overall \\
\midrule
\agentlogo{anthropic}\href{https://artificialanalysis.ai/models/claude-opus-5-high}{Claude Opus 5} & 52 & 65.41 \\
\agentlogo{xai}\href{https://artificialanalysis.ai/models/grok-4-6}{Grok 4.6} & 51 & 62.93 \\
\agentlogo{openai}\href{https://artificialanalysis.ai/models/gpt-5-6-sol-high}{GPT-5.6 Sol} & 48 & 57.28 \\
\agentlogo{gemini}\href{https://artificialanalysis.ai/models/gemini-3-8-flash}{Gemini 3.8 Flash} & 47 & 59.57 \\
\bottomrule
\end{tabular}
\end{table}

As shown in Table~\ref{tab:aa-correlations}, our benchmark's Overall score exhibits a strong positive rank correlation ($\rho=0.800$) with the AA Intelligence Index, while Activation Selectivity achieves a perfect rank correlation ($\rho=1.000$). These results indicate that agentic mechanistic interpretability capabilities align broadly with general frontier model intelligence, while retaining distinct evaluative variance in causal steering and dictionary-wide ranking.

\begin{table}[H]
\caption{Spearman rank correlation of benchmark scores with the AA Intelligence Index.}
\label{tab:aa-correlations}
\centering\small\setlength{\tabcolsep}{14pt}
\begin{tabular}{@{}lr@{}}
\toprule
Benchmark dimension & Spearman $\rho$ \\
\midrule
Activation Selectivity & 1.000 \\
Overall Score & 0.800 \\
Activation Rank & 0.400 \\
Causal Steering & 0.400 \\
\bottomrule
\end{tabular}
\end{table}

\subsection{Performance Across Tasks and Categories}
\label{app:additional-agent-analysis}

\begin{figure}[H]
\centering\includegraphics[width=\linewidth]{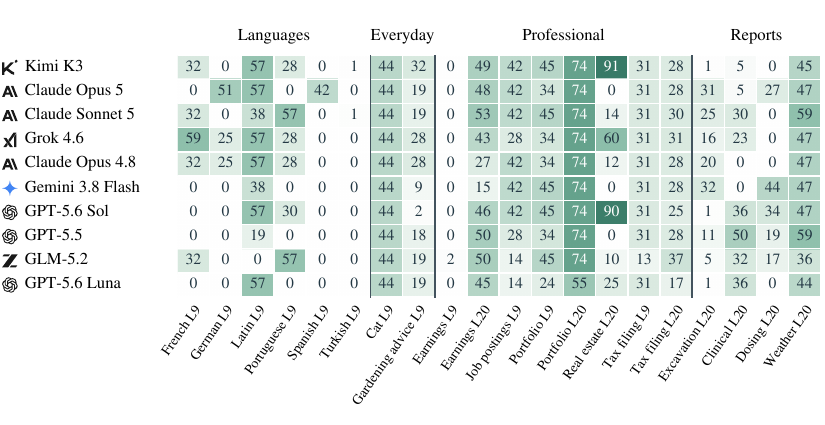}
\caption{Mean Steering for every agent and task. Values are rounded to integers; color spans 0--100. L9 and L20 identify the SAE layer.}
\label{fig:steering-all-tasks}
\end{figure}

\paragraph{Steering by task.}
Figure~\ref{fig:steering-all-tasks} reports each agent's mean Steering on every task, averaging its three investigations. Shared columns identify common outcomes, while differences within a column compare agents on the same target and layer. Kimi and Sol obtain positive real-estate Steering scores, whereas Opus 5 scores zero on that task. Opus 5 instead obtains nonzero Spanish Steering, where most agents score zero. Cat has the same result for every agent. The matrix makes these task-specific differences visible alongside the overall ranking.

\begin{figure}[H]
\centering\includegraphics[width=\linewidth]{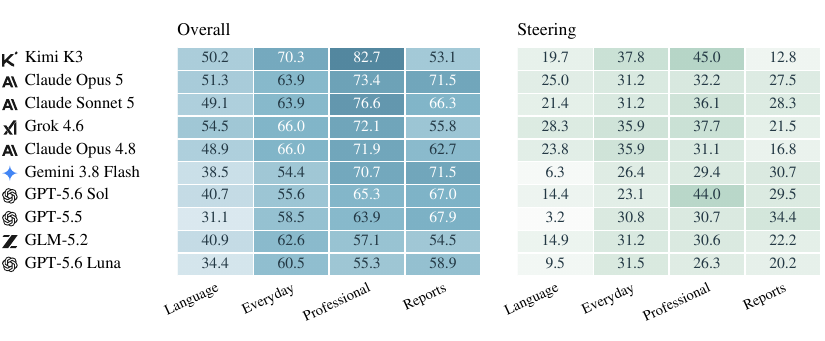}
\caption{Overall and Steering within each task category. Cells show the adopted scores, with a common 0--100 color range for the displayed values.}
\label{fig:category-scores-all}
\end{figure}

\paragraph{Comparing categories.}
Figure~\ref{fig:category-scores-all} groups the same measurements into six language, two everyday, eight professional, and four specialized-report tasks. Each category weights its tasks equally. The benchmark's Overall score weights all 20 tasks equally, so these four category means contribute in proportion to their task counts. Kimi's professional Steering of 45.03 exceeds Opus 5's 32.19, while Opus leads Kimi on reports, 27.50 versus 12.81. Sol is close to Kimi on professional Steering at 44.04, despite a lower category Overall of 65.30 versus 82.71. These differences link the component ranking to the tasks on which each agent obtains its scores.

\paragraph{Repeatability across all tasks.}
Table~\ref{tab:repeatability-all} reports identical selections and the advantage of taking the retrospectively best investigation for each task. Together with the task matrix, it distinguishes consistently repeated choices from the feature with the highest Steering score across three attempts.
\begin{table}[H]
\caption{Repeated feature selections and mean versus best Steering across three runs on 20 tasks.}
\label{tab:repeatability-all}
\centering\small\setlength{\tabcolsep}{6pt}\renewcommand{\arraystretch}{1.02}
\begin{tabular}{@{}lrrrr@{}}
\toprule
 & Same choice & \multicolumn{3}{c}{Steering} \\
\cmidrule(l){3-5}
Agent & across runs & Mean & Best & Gap \\
\midrule
\agentlogo{kimi}Kimi K3 & 50\% & 30.26 & 38.09 & 7.83 \\
\agentlogo{anthropic}Claude Opus 5 & 45\% & 28.99 & 37.00 & 8.01 \\
\agentlogo{anthropic}Claude Sonnet 5 & 35\% & 29.66 & 42.94 & 13.28 \\
\agentlogo{xai}Grok 4.6 & 35\% & 31.47 & 46.00 & 14.53 \\
\agentlogo{anthropic}Claude Opus 4.8 & 45\% & 26.55 & 38.34 & 11.79 \\
\agentlogo{gemini}Gemini 3.8 Flash & 55\% & 22.43 & 26.66 & 4.23 \\
\agentlogo{openai}GPT-5.6 Sol & 50\% & 30.16 & 37.69 & 7.53 \\
\agentlogo{openai}GPT-5.5 & 30\% & 23.19 & 32.31 & 9.12 \\
\agentlogo{zai}GLM-5.2 & 20\% & 24.28 & 38.16 & 13.88 \\
\agentlogo{openai}GPT-5.6 Luna & 30\% & 20.54 & 29.38 & 8.83 \\
\bottomrule
\end{tabular}

\end{table}

\FloatBarrier
\subsection{Repeated Discoveries on Four Concepts}
\begin{figure}[H]
\centering\includegraphics[width=\linewidth]{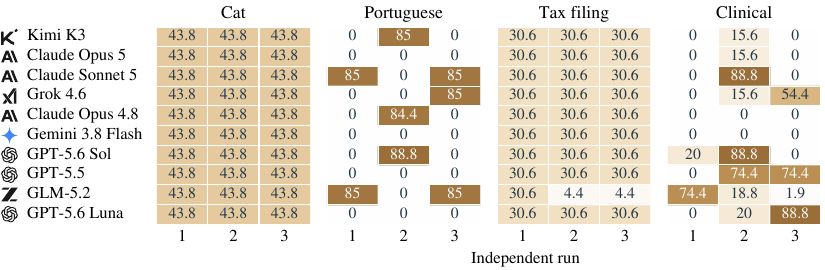}
\caption{Steering scores across independent discoveries of four concepts.}
\label{fig:cross-agent-runs}
\end{figure}

\end{document}